\documentclass{article}

\PassOptionsToPackage{numbers, compress}{natbib}

\usepackage{xcolor}         
\usepackage[main,final]{neurips_2026}
\usepackage{subcaption}

\usepackage[utf8]{inputenc} 
\usepackage[T1]{fontenc}    

\usepackage{graphicx}
\usepackage{url}
\usepackage{hyperref}
\usepackage{xspace}
\usepackage{microtype}
\usepackage{nicefrac}

\usepackage{amsmath}
\usepackage{amssymb}
\usepackage{amsfonts}

\usepackage[table]{xcolor}
\usepackage{booktabs}
\usepackage{multirow}
\usepackage{makecell}
\usepackage{array}
\usepackage{pifont}

\usepackage{wrapfig}
\usepackage{caption}

\usepackage{algorithm}
\usepackage{algorithmic}

\usepackage{tabularx}
\usepackage[table]{xcolor}
\usepackage{booktabs}
\usepackage{amsmath}
\usepackage{amssymb}
\definecolor{orange}{RGB}{255,140,0}

\newcommand{\cls}{\texttt{[CLS]}\xspace}

\title{Lift, See, Act: Hierarchical Robot Policy Pretraining with 3D Foundation Models}

\author{%
  \textbf{Yiyuan Ge} \quad
  \textbf{Changxing Ding}\thanks{Corresponding author.} \quad
  \textbf{Ziyu Hao} \quad
  \textbf{Zijie Zheng} \quad
  \textbf{Xiangmin Xu} \\
  School of Electronic and Information Engineering \\
South China University of Technology \\
  \texttt{\{202510182871,202610183440,202620112031\}@mail.scut.edu.cn} \\
\texttt{\{chxding,xmxu\}@scut.edu.cn}
}

\begin{document}

\maketitle

\begin{abstract}


Robot policy pretraining based on human videos is crucial for improving the policy’s generalization ability. One main challenge for this task is the lack of explicit action-relevant representations in such unlabeled data. Recent works tend to estimate 3D hand motion trajectories from videos using 3D foundation models (3D FMs). However, they keep solely the sparse trajectories as supervision, ignoring the rich, fine-grained information produced during the trajectory extraction process. To address this issue, we propose \textbf{LSA}, a hierarchical robot policy pretraining framework that includes three phases: \textbf{L}ift, \textbf{S}ee, and \textbf{A}ct. Specifically, the Lift phase elevates 2D video observations into dense 3D representations by adaptively aligning features in the policy's shallow layers with the intermediate representations extracted from multiple 3D FMs. Building upon these lifted representations, the See phase equips the policy with explicit geometric and interactive awareness through dual guidance. It introduces depth-map reconstruction to help the model comprehend 3D spatial layouts and utilizes hand region cues to explicitly supervise the encoder’s attention toward task-relevant interaction hotspots. Empowered by the dense 3D representation learning and precise spatial guidance, our policy achieves robust and accurate robotic manipulation in the Act phase. Extensive experiments on diverse simulation and real-world tasks demonstrate that LSA significantly outperforms current state-of-the-art approaches. The project page is available at:
\href{https://yiyuan00.github.io/Lift-See-Act}{\textcolor{blue}{https://yiyuan00.github.io/Lift-See-Act}}.
\end{abstract}

\section{Introduction}
\label{sec:intro}

Large-scale pretraining is crucial for improving the generalization ability of machine learning models \cite{gpt,llava}. In robotics, pretraining is equally important for policy learning, since high-quality manipulation data are expensive to collect and therefore difficult to scale up. Human videos are a key source of pretraining data, as they cover a wide range of real-world manipulation scenarios and contain rich manipulation knowledge \cite{clap,dreamdojo}. However, using human videos to pretrain robot policies remains a challenge, as such data inherently lack action-relevant representations \cite{survey}.

Existing methods for extracting explicit action-relevant representations from human videos can be divided into two categories. The first category of methods extracts 2D 
visual primitives from video frames, such as contact points \cite{maple, versa}, contact heatmaps \cite{hrp}, and object bounding boxes \cite{mpi}, to characterize where contact occurs and which regions are potentially involved in manipulation. However, these methods lack temporal representations of the 3D manipulation process. The second category of methods attempts to estimate 3D hand motion trajectories from videos \cite{beingh0,vidbot,vitra,deximit,videomanip}. These trajectories reflect both the displacement of the hand in the 3D space and the temporal evolution of that displacement; therefore, they are more in line with the robot’s action prediction space. Recently, with the development of 3D foundation models (3D FMs) \cite{vggt,dpro,moge2,hawor}, the accuracy of the estimation of 3D hand motion trajectories has improved substantially \cite{vitra,deximit,videomanip}.
\begin{wrapfigure}{r}{0.48\textwidth} 
  \begin{minipage}{\linewidth}
    \centering
    \includegraphics[width=\linewidth]{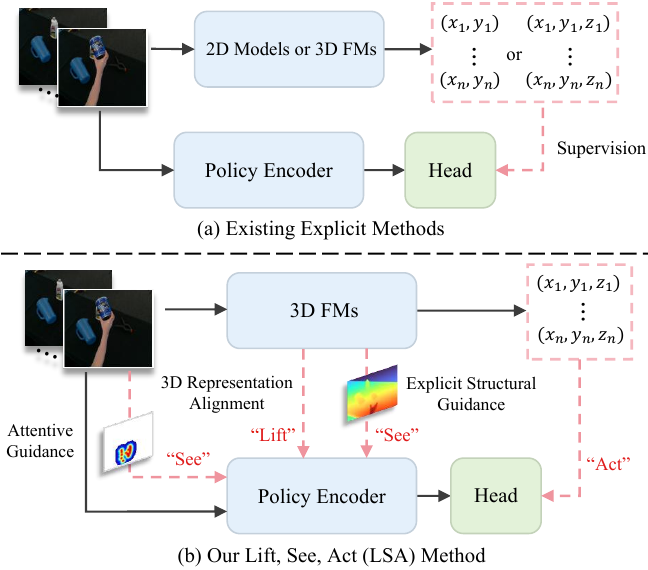} 
    \vspace{-1em} 
    \caption{
    Comparisons between methods that extract explicit action-relevant representations from human videos. (a) Existing methods adopt solely the estimated hand motion trajectories as supervision~\cite{hrp,vitra,beingh0}. (b) Our LSA leverages multi-tiered information from multiple 3D FMs to facilitate robot policy pretraining.} 
    \label{fig:f1} 
  \end{minipage}
  \vspace{-2em} 
\end{wrapfigure}
However, as illustrated in Fig.~\ref{fig:f1}, the latter category of approaches only keeps the estimated hand motion trajectories as supervision for robot policy pretraining. They ignore more fine-grained knowledge, e.g., rich geometric representations and depth maps, produced by the 3D FMs during the hand motion trajectory prediction process. This dense knowledge is fundamental for a robot policy in predicting the 3D motion of the end effector. This insight motivates us to study how to make the best use of multi-tiered knowledge from 3D FMs to promote robot policy pretraining.

Consequently, we propose \textbf{LSA}, a hierarchical framework for robot policy pretraining based on three phases: \textbf{Lift}, \textbf{See}, and \textbf{Act}. Specifically, in the \textbf{Lift} phase, our goal is to lift 2D video observations into rich 3D representations. We first construct a 3D hand motion trajectory extraction pipeline utilizing multiple 3D FMs to process unlabeled human videos. This pipeline not only extracts the motion trajectories, but also keeps dense intermediate representations and depth maps produced by these 3D FMs. We then enforce the shallow-layer representations of the policy to align with the intermediate representations of each 3D FM, respectively. We also propose a strategy to adaptively adjust the weights of these alignment losses. This process effectively lifts the policy's 2D representations into 3D, endowing the policy with more fine-grained geometric knowledge.

Building upon these lifted representations, the \textbf{See} phase equips the policy with explicit geometric and interactive awareness. While the Lift phase establishes a dense 3D representation space, these representations are implicit. Robot manipulation inherently requires an intuitive and precise understanding of the structure of the physical scene, along with a distinct visual focus on the manipulation objectives. To achieve this, the See phase provides dual guidance: explicit structural guidance and attentive guidance. First, leveraging the extracted depth maps, we introduce depth map reconstruction as an auxiliary task, providing explicit 3D geometric constraints to enable the model to comprehend 3D spatial layouts. Second, we utilize the hand region as a cue to guide the encoder's attention toward task-relevant interaction hotspots. 
Specifically, we encode 2D hand keypoints into a Gaussian heatmap to represent the hand region. We then select the attention map most similar to this heatmap across all heads in the shallow layers as our target, and finally use it to supervise the mean attention of all heads within a layer.
The above dual guidance ensures the policy explicitly \textbf{Sees} the 3D scene structure and concentrates on the task-relevant interaction regions.

To the best of our knowledge, LSA is the first approach to leverage multi-tiered 3D information extracted from 3D foundation models for hierarchical robot policy pretraining on human videos. Empowered by the dense 3D representation learning in the Lift phase and the dual guidance in the See phase, our policy predicts 3D hand motion trajectories in the \textbf{Act} phase, ultimately achieving robust and accurate robot manipulation. Extensive experiments demonstrate the effectiveness of our design and show that LSA outperforms state-of-the-art methods on various simulation and real-world tasks.

\section{Related Works}
\label{sec:rela}

Existing approaches to pretraining robot policies from human videos often construct action-relevant representations to bridge the gap between passive visual observations and robot control. Based on the space in which these representations are extracted, existing approaches can be broadly categorized into 2D and 3D methods.

\textbf{2D Methods.} The first category of methods constructs action-relevant representations within the 2D image space, which can be further divided into implicit and explicit methods. 
Implicit methods model inter-frame relationships to extract abstract state representations. One line of work uses contrastive learning \cite{r3m,vip} to constrain the distance between visual embeddings across different time periods, capturing the temporal dynamics of visual observations (i.e., how the current state transitions to the next). To characterize these state transitions more precisely, other approaches learn discrete latent actions between frames through VQ-VAE objectives \cite{vqvae}, subsequently using these latent actions as pseudo-action labels for unannotated videos \cite{clap,lapa,como}. Recently, some methods train video prediction models \cite{vpp,gen2act,vipra,dreamdojo} to extract predictive visual representations, providing embodied agents with dense information about future states. 

In contrast, explicit methods directly extract 2D visual primitives at the pixel level. These approaches typically identify contact points \cite{maple,versa}, contact heatmaps \cite{hrp}, or bounding boxes \cite{mpi,precise} to delineate potential interaction regions. Furthermore, extracting human hand poses from video streams serves as a simple but scalable approach to represent the interaction trajectories \cite{versa,mas}.

\textbf{3D Methods.} The above 2D representations fundamentally lack explicit depth information, limiting their ability to capture inherently 3D physical interactions. To address this limitation, the second category of methods turns to explicit 3D action-relevant representations, such as 3D hand trajectories \cite{vitra} and parametric hand poses (e.g., MANO \cite{mano}) \cite{egovla,maple}. Due to their 3D nature, these representations align more tightly with the robotic action space. 

These 3D methods can be further divided into two groups according to their data requirements. The first group uses videos equipped with annotated 3D hand poses and camera parameters, typically captured through AR/VR devices or RGB-D SLAM~\cite{slam}. 
They directly adopt these ground-truth 3D hand poses as supervision for the pretraining of robot policies \cite{egovla,beingh0,hrdt,innon}.
However, these videos are recorded in relatively controlled settings, which affects the scale of the video databases. In comparison, the second group constructs 3D representations directly from unlabeled, in-the-wild videos. Early efforts employ structure-from-motion and depth estimation models to recover camera intrinsics and metric depth. They then combine these estimations with continuous 2D hand poses to obtain 3D hand motion trajectories \cite{vidbot, develop}. Recently, the rapid advancement of 3D Foundation Models (FMs) \cite{vggt,moge2,dpro} has enabled the extraction of 3D hand trajectories with substantially improved fidelity and efficiency \cite{vitra, beingh0}. 

In this paper, we also use 3D FMs to construct action-relevant representations. Unlike trajectory-based methods that focus exclusively on 3D hand trajectories and Spatial Forcing~\cite{sforcing}, which distills dense geometric features from a single 3D FM, we extract hierarchical representations (e.g., geometric features and depth maps) from multiple 3D FMs during inference, providing multi-level 3D supervision for robot policy pretraining.

\section{Methods}
\label{sec:method}
\subsection{Problem Formulation and LSA Overview}
\label{sec:pre}
\begin{figure*}[t]
  \centering
  \includegraphics[width=\linewidth]{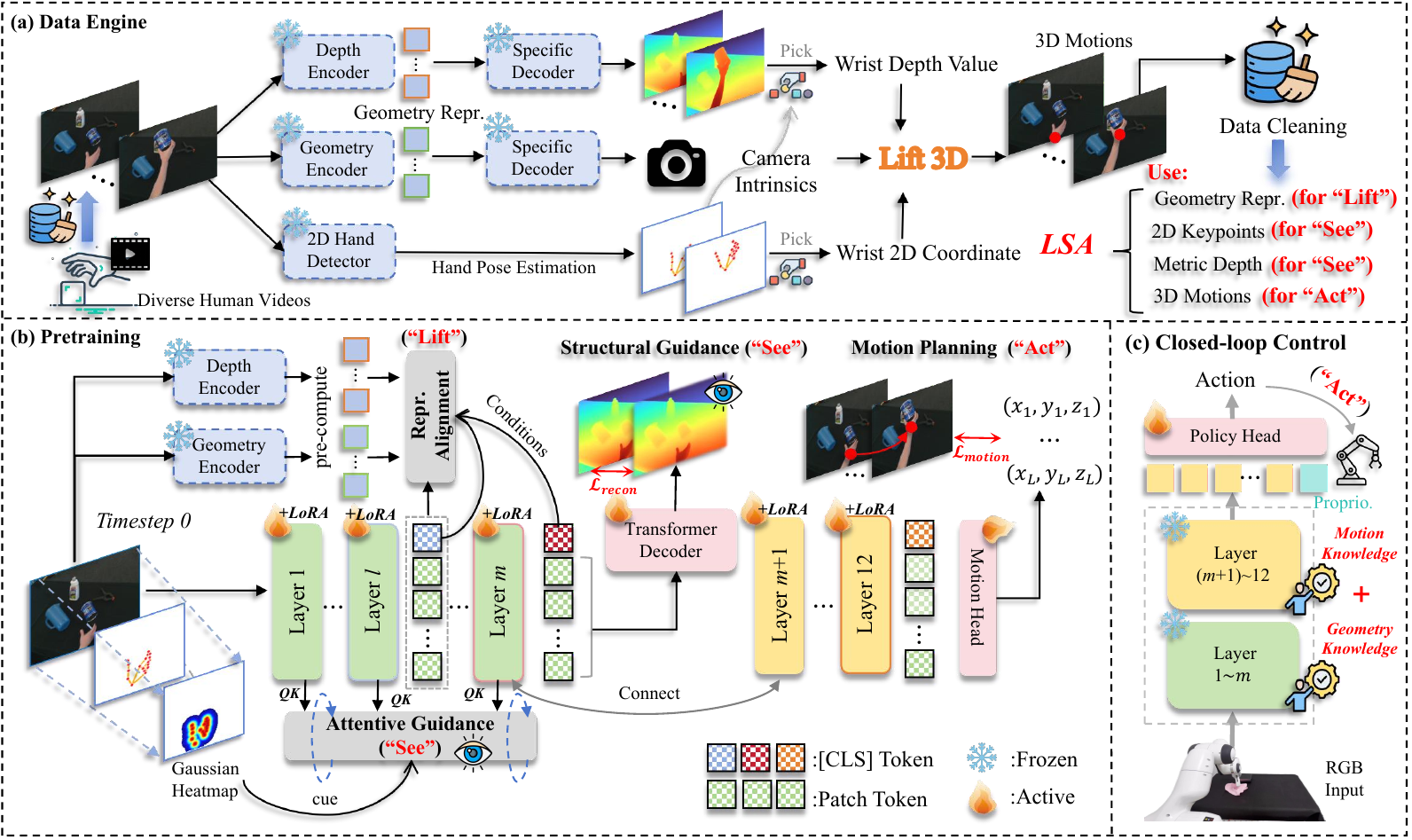}
  \caption{ 
  Architecture overview of LSA. (a) In the data engine phase, we utilize two 3D FMs to extract depth maps and camera intrinsics, respectively. Combined with the extracted 2D wrist coordinates, we obtain 3D hand motion trajectories. Furthermore, we also extract dense geometric representations and depth maps. (b) In the pretraining phase, we first align the policy's shallow-layer representations with the intermediate representations of the two 3D FMs pre-computed during the data engine phase. Then, we introduce depth-map reconstruction to equip the model with precise structural knowledge, and utilize hand heatmaps to guide the model's attention toward interaction hotspots. Following this, we predict the 3D hand motion trajectories. (c) Finally, we freeze the pretrained LSA and combine it with a policy head to perform closed-loop robotic control.}
  \label{fig:framework}
\end{figure*}

Following prior work \cite{hrp,r3m,maple,maskworld,miti}, we pretrain the visual encoder part of the robot policy using large-scale human videos. 
Once trained, the visual encoder $F_{\phi^*}$, parameterized by the fixed pretrained parameters $\phi^*$, is frozen and serves as the perceptual front-end of the robot policy. Given a current visual observation $\mathbf{I}$, the encoder maps it to a latent embedding $\mathbf{z}=F_{\phi^*}(\mathbf{I})$.
This embedding $\mathbf{z}$ is then concatenated with the proprioceptive state of the robot $\mathbf{s}$ and fed into a policy head $\pi_{\theta}$, parameterized by $\theta$, to generate an action $\hat{\mathbf{a}} \sim \pi_{\theta}(\cdot \mid \mathbf{z}, \mathbf{s})$. The policy is trained through behavior cloning \cite{bc} using a dataset of expert demonstration trajectories $\mathcal{D}_{\mathrm{IL}} = \{ \tau_i \}_{i=1}^{N_{\mathrm{demo}}}$, where each trajectory is defined as $\tau_i=\left((\mathbf{I}_{i,t},\mathbf{s}_{i,t},\mathbf{a}_{i,t})\right)_{t=1}^{T}$. Here, $N_{\mathrm{demo}}$ is the number of demonstration trajectories and $T$ is the trajectory length. The training objective minimizes the behavior cloning loss between the predicted action $\hat{\mathbf{a}}_{i,t}$ and the ground-truth expert action $\mathbf{a}_{i,t}$:
\begin{equation}
\min_{\theta}
\mathbb{E}_{\tau_i\sim\mathcal{D}_{\mathrm{IL}},\,
t\sim\mathcal{U}\{1,\ldots,T\}}
\left[
\mathcal{L}_{\mathrm{BC}}
\left(
\hat{\mathbf{a}}_{i,t},
\mathbf{a}_{i,t}
\right)
\right].
\end{equation}
Here, $\mathcal{L}_{\mathrm{BC}}$ denotes the behavior cloning loss, instantiated in our implementation as the negative log-likelihood of the expert action under the predicted policy distribution. During evaluation, the policy is deployed in a closed-loop control setting and its performance is quantified by the success rate averaged over multiple deployments. 

As highlighted in Section \ref{sec:intro}, relying solely on estimated hand trajectories for robot policy pretraining wastes the fine-grained geometric knowledge inherently produced by 3D Foundation Models (3D FMs). To make the best use of this multi-tier knowledge, we propose \textbf{LSA}, a hierarchical pretraining framework consisting of three distinct phases: \textbf{Lift}, \textbf{See}, and \textbf{Act}. 
First, the \textbf{Lift} phase elevates 2D video observations into a dense 3D representation space by aligning the policy's shallow-layer features with intermediate representations from 3D FMs. Building upon this, the \textbf{See} phase equips the policy with explicit geometric and interactive awareness through structural and attentive dual guidance. Finally, empowered by these representations, the \textbf{Act} phase translates visual inputs into 3D hand motion trajectories for robust manipulation. 

\subsection{The Lift Phase: 3D Representation Elevation}

\textbf{Data Engine.}
To extract 3D hand motion trajectories, we develop a data engine using multiple 3D FMs. Given an input video frame, we first extract the 2D wrist coordinates $\mathbf{p}_w$. Subsequently, we employ VGGT \cite{vggt} to estimate camera intrinsics $\mathbf{K}$, and Depth Pro \cite{dpro} to predict a metric depth map $\mathbf{D}_g$. 
The intermediate dense features extracted from VGGT and Depth Pro are denoted by $\mathbf{E}_c$ and $\mathbf{E}_d$, respectively.
Using $\mathbf{K}$ and $\mathbf{D}_g$, we lift the 2D wrist coordinates $\mathbf{p}_w$ into the 3D space. Finally, we conduct necessary data cleaning to obtain high-quality pretraining data. Details of data cleaning and 3D lifting are provided in \textbf{Appendix~\ref{app:data_cleaning}}.

\textbf{Dense Representation Alignment.}
Geometric representations from 3D FMs have been shown to benefit robot policy learning~\cite{sforcing}. Given an input image $\mathbf{I}_h \in \mathbb{R}^{H \times W \times 3}$, the vision encoder extracts visual features from it. Let $\mathbf{X}_a \in \mathbb{R}^{N_a \times D_a}$ denote the visual tokens from the $l$-th layer of the encoder. To distill the fine-grained geometric knowledge of the two 3D FMs into the policy, we align the shallow-layer visual tokens $\mathbf{X}_a$ with their geometric representations $\mathbf{E}_c \in \mathbb{R}^{N_c \times D_c}$ and $\mathbf{E}_d \in \mathbb{R}^{N_d \times D_d}$, respectively.

To ensure spatial resolution compatibility, we first reduce the number of tokens in $\mathbf{E}_c$ and $\mathbf{E}_d$ to match $N_a$ via downsampling and add positional encodings to their respective image patch tokens. We denote the resulting spatially-aligned features as $\mathbf{E}'_c \in \mathbb{R}^{N_a \times D_c}$ and $\mathbf{E}'_d \in \mathbb{R}^{N_a \times D_d}$. 
Next, to ensure compatibility with the feature dimensions, we map the embedding dimension of $\mathbf{X}_a$ to $D_c$ and $D_d$ using two separate MLPs, respectively. Following the typical token-level alignment paradigm \cite{repa}, we employ a cosine similarity score to maximize the alignment between the visual tokens and the geometric signals. To balance the contributions of multiple geometric features, we introduce a set of dynamic modulation weights $w_c$ and $w_d$. The overall alignment loss is formulated as:
\begin{equation}
    \label{eq:alignment}
    \mathcal{L}_{\mathrm{align}} = - \frac{1}{N_a} \sum_{i=1}^{N_a} \left[ w_c \cdot \mathcal{S}(\mathrm{MLP}_c(\mathbf{x}_{a,i}), \mathbf{e}'_{c,i}) + w_d \cdot \mathcal{S}(\mathrm{MLP}_d(\mathbf{x}_{a,i}), \mathbf{e}'_{d,i}) \right],
\end{equation}
where $\mathbf{x}_{a,i}$ denotes the $i$-th visual token in $\mathbf{X}_a$, and $\mathbf{e}'_{c,i}$ and $\mathbf{e}'_{d,i}$ represent the tokens in $\mathbf{E}'_c$ and $\mathbf{E}'_d$ that spatially correspond to $\mathbf{x}_{a,i}$, respectively.
$\mathcal{S}(\cdot, \cdot)$ denotes the cosine similarity function. This alignment process effectively lifts the policy's 2D representations into 3D, endowing it with fine-grained geometric awareness.

To ensure that the model captures valuable geometric knowledge from multiple teachers, we propose a dynamic weight assignment mechanism inspired by~\cite{movekd}. We derive a condition query by aggregating the \cls tokens $\mathbf{c}_l$ and $\mathbf{c}_m$ from the $l$-th layer and the $m$-th layer of the vision encoder, where the $m$-th layer is further used for depth reconstruction. Then, we compute the scaled dot-product similarity between this query and the geometric tokens to adaptively determine the dynamic weights $w_c$ and $w_d$ for $\mathbf{E}'_c$ and $\mathbf{E}'_d$, respectively. More details are provided in \textbf{Appendix}~\ref{app:dynamic_weight}.


\subsection{The See Phase: Dual Attentive and Structural Guidance}
Although the Lift phase establishes a dense 3D representation space, these features remain largely implicit. Robot manipulation inherently requires an intuitive understanding of the structure of the physical scene, alongside a distinct visual focus on the manipulating agent and the target object. To achieve this, the See phase provides dual guidance.

\textbf{Attentive Guidance.}
\begin{figure*}[t]
  \centering
  \includegraphics[width=0.9\linewidth]{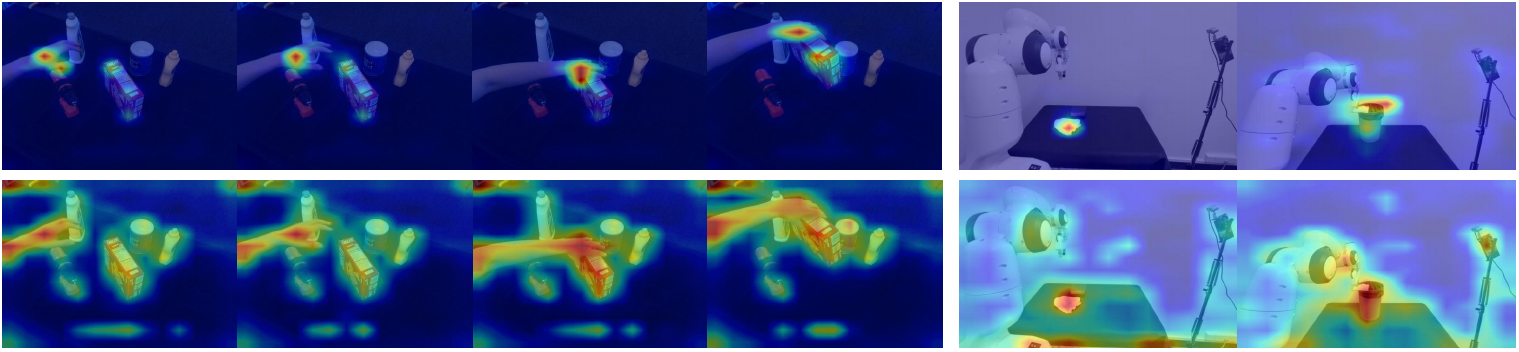}
  \caption{
Effects of attentive guidance (AG). Left: attention maps on human videos. Right: attention maps on robotic scenes. In both cases, the top row shows results w/ AG and the bottom row shows results w/o AG.
  }
  \vspace{-1em}
  \label{fig:ag_vis}
\end{figure*}
Learning affordance knowledge from human videos benefits policy pretraining~\cite{hrp,maple}. To further improve the policy's perception of hand-centric interaction regions, we design an attentive guidance mechanism. Specifically, we first warm up the LSA visual encoder for several epochs with the motion planning objective, allowing initial task-relevant attention patterns to emerge (see \textbf{Appendix}~\ref{app:att_guidance}). Given the detected hand keypoints, we generate a Gaussian heatmap and downsample it to the patch resolution, obtaining a spatial hand prior $\mathbf{h} \in \mathbb{R}^{N_a}$, where $N_a$ denotes the number of spatial patches. For each head in the first $R$ layers, we extract the $\mathrm{[CLS]}$-to-patch attention map $\mathbf{A}_{n,k} \in \mathbb{R}^{N_a}$, where $n$ and $k$ denote the layer and head indices. We then compute its alignment score with the hand prior as:
\begin{equation}
    \label{eq:relevance}
    s_{n,k} = \frac{1}{N_a}\sum_{i=1}^{N_a} A_{n,k,i}h_i .
\end{equation}
The attention map with the highest score is selected as the guidance target:
\begin{equation}
    \label{eq:selection}
    (n^*, k^*) = \arg\max_{n,k} s_{n,k}, 
    \qquad 
    \mathbf{A}^* = \mathbf{A}_{n^*,k^*}.
\end{equation}
We then use $\mathbf{A}^*$ as the self-distillation target to refine the averaged shallow attention. For each layer $n \leq R$, we compute the averaged multi-head attention map $\bar{\mathbf{A}}_n$ and minimize its distance to $\mathbf{A}^*$ after spatial normalization:
\begin{equation}
    \label{eq:att_loss}
    \mathcal{L}_{\mathrm{att}}
    =
    \sum_{n=1}^{R}
    \left\|
    \mathcal{N}(\mathrm{sg}(\mathbf{A}^*))
    -
    \mathcal{N}(\bar{\mathbf{A}}_n)
    \right\|_2^2 ,
\end{equation}
where $\mathcal{N}(\cdot)$ denotes Min-Max normalization and $\mathrm{sg}(\cdot)$ denotes stop-gradient. In this way, attentive guidance transfers the model's intrinsic hand-centric attention pattern to the shallow layers, enabling them to better focus on task-relevant hand-object regions.

As shown in Fig.~\ref{fig:ag_vis}, we visualize the averaged attention maps over the first $R$ shallow layers to analyze the effect of attentive guidance.
After applying attentive guidance, the model shows a clearer focus on the hand region.
Notably, it also exhibits emergent attention to interacting objects, which can be attributed to the motion planning (\textbf{Sec.}~\ref{sec:mp}) objective  that tightly couples future hand trajectories with hand-object spatial relations.
Moreover, Fig.~\ref{fig:ag_vis} shows that these refined attention patterns transfer to robotic manipulation scenes, enabling the model to focus more on task-relevant regions while reducing attention to background.


\textbf{Explicit Structural Guidance.}
To provide explicit 3D geometric constraints and enable the model to comprehend 3D spatial layouts more precisely, we introduce depth-map reconstruction as an auxiliary task. We utilize the image patch tokens from the $m$-th layer of the encoder, denoted as $\mathbf{X}_m$, which are processed by a transformer decoder to regress the depth map. This reconstruction is supervised by the metric depth map $\mathbf{D}_g$ generated during the data engine stage, and the reconstruction loss $\mathcal{L}_{\mathrm{recon}}$ is computed using mean squared error (MSE).


\subsection{The Act Phase: Motion Planning}
\label{sec:mp}
Empowered by the dense 3D representation learning in the Lift phase and the precise dual guidance in the See phase, the Act phase predicts the 3D hand motion trajectory. For a training sample with $L+1$ sequential frames, we first lift the wrist locations into 3D space using the estimated camera intrinsics and the metric depth map, and construct the ground-truth future trajectory $\mathbf{T} \in \mathbb{R}^{L \times 3}$ from the subsequent $L$ frames. We then average-pool the final-layer visual tokens of the initial frame into a global visual embedding, which is fed into a single-layer MLP prediction head to predict the future trajectory $\hat{\mathbf{T}} \in \mathbb{R}^{L \times 3}$. Finally, the motion planning objective $\mathcal{L}_{\mathrm{motion}}$ is optimized by minimizing the $\ell_1$ distance between $\hat{\mathbf{T}}$ and $\mathbf{T}$.
By grounding action prediction in these rich 3D priors and refined attention, the policy acquires robust, generalized motion planning capabilities, establishing a strong foundation for downstream robot control.


\section{Simulation Experiments}
\label{sec:sim_exp}

\subsection{Setup}
\label{sec:sim_setup}

\textbf{Evaluation Setup.}
We evaluate LSA on three simulation benchmarks: MetaWorld~\cite{meta}, DexArt~\cite{dexart}, and Adroit~\cite{adroit}.
MetaWorld evaluates tabletop manipulation with a Sawyer arm and gripper, while DexArt and Adroit focus on dexterous manipulation with multi-fingered hands.
For MetaWorld, we use eight tasks covering different difficulty levels: \textit{hammer}, \textit{dial turn}, \textit{assembly}, \textit{hand insert}, \textit{pick place wall}, \textit{shelf place}, \textit{disassemble}, and \textit{stick pull}.
For DexArt, we evaluate on \textit{bucket} and \textit{toilet}; for Adroit, we evaluate on \textit{door}, \textit{hammer}, and \textit{pen}.
Following prior methods, the pretrained visual encoder is frozen and combined with a lightweight behavior cloning policy head.
We report success rates averaged over three random seeds.
Detailed data collection, training, and evaluation protocols are provided in \textbf{Appendix~\ref{app:sim_setup}}.

\textbf{Baselines.}
We compare LSA with three groups of baselines.
First, we include general-purpose visual representations, including CLIP~\cite{clip} and DINOv3~\cite{dinov3}.
Second, we compare with representative robot visual representation pretraining methods, including MCR~\cite{mcr}, VIP~\cite{li2024vip}, R3M~\cite{r3m}, VC-1~\cite{vc1}, HRP~\cite{hrp}, MPI~\cite{mpi}, and R3M-A~\cite{miti}.
Third, we include baselines that directly take 3D point clouds as inputs, including DP3~\cite{dp} and KAN-We-Flow~\cite{kwf}.
For VIP, we additionally provide cropped target images following its original protocol.
More baseline details are provided in \textbf{Appendix~\ref{app:baseline_details}}.

\subsection{Results}
\label{sec:results}

\begin{table*}[t]
\centering
\caption{Success rate (\%) reported as mean $\pm$ standard deviation over 3 random seeds on MetaWorld, DexArt, and Adroit. $^\ast$ denotes that cropped target images are used as additional inputs. Task abbreviations: HA (Hammer), DT (Dial Turn), AS (Assembly), HI (Hand Insert), PPW (Pick Place Wall), ShP (Shelf Place), DI (Disassemble), StP (Stick Pull), BU (Bucket), TO (Toilet), DR (Door), HA-A (Adroit Hammer), and PE (Pen). Bold and underline denote the best and second-best results, respectively.}
\label{tab:sim_results}
\tiny
\setlength{\tabcolsep}{1.15pt}
\renewcommand{\arraystretch}{1.05}
\resizebox{\textwidth}{!}{
\begin{tabular}{lccccccccccccc}
\toprule
\textbf{Method} 
& \multicolumn{8}{c}{\textbf{MetaWorld}} 
& \multicolumn{2}{c}{\textbf{DexArt}} 
& \multicolumn{3}{c}{\textbf{Adroit}} \\
\cmidrule(lr){2-9} \cmidrule(lr){10-11} \cmidrule(lr){12-14}
& HA & DT & AS & HI & PPW & ShP & DI & StP & BU & TO & DR & HA-A & PE \\
\midrule
\multicolumn{14}{l}{\textit{Direct 3D Policy}} \\
DP3 (\textit{RSS'24}) 
& $\text{91.1}_{\textcolor{gray}{\pm\text{2.8}}}$ 
& $\text{84.9}_{\textcolor{gray}{\pm\text{2.0}}}$ 
& $\text{51.1}_{\textcolor{gray}{\pm\text{3.3}}}$ 
& $\text{33.8}_{\textcolor{gray}{\pm\text{4.0}}}$ 
& $\text{32.9}_{\textcolor{gray}{\pm\text{3.0}}}$ 
& $\text{48.4}_{\textcolor{gray}{\pm\text{4.7}}}$ 
& $\text{40.4}_{\textcolor{gray}{\pm\text{3.5}}}$ 
& $\text{39.1}_{\textcolor{gray}{\pm\text{4.0}}}$ 
& $\text{26.7}_{\textcolor{gray}{\pm\text{1.7}}}$ 
& $\text{71.7}_{\textcolor{gray}{\pm\text{2.2}}}$ 
& $\text{63.3}_{\textcolor{gray}{\pm\text{1.7}}}$ 
& $\textbf{100.0}_{\textcolor{gray}{\pm\text{0.0}}}$ 
& $\text{40.6}_{\textcolor{gray}{\pm\text{1.0}}}$ \\

KAN-We-Flow (\textit{ICRA'26}) 
& $\text{89.3}_{\textcolor{gray}{\pm\text{1.4}}}$
& $\text{89.8}_{\textcolor{gray}{\pm\text{2.8}}}$
& $\text{61.8}_{\textcolor{gray}{\pm\text{4.1}}}$
& $\text{58.7}_{\textcolor{gray}{\pm\text{1.4}}}$
& $\text{68.0}_{\textcolor{gray}{\pm\text{4.0}}}$
& $\text{70.2}_{\textcolor{gray}{\pm\text{2.8}}}$
& $\text{68.9}_{\textcolor{gray}{\pm\text{2.0}}}$
& $\text{66.7}_{\textcolor{gray}{\pm\text{4.0}}}$
& $\text{31.1}_{\textcolor{gray}{\pm\text{2.6}}}$
& $\underline{\text{76.7}}_{\textcolor{gray}{\pm\text{1.7}}}$
& $\underline{\text{80.5}}_{\textcolor{gray}{\pm\text{2.5}}}$
& $\textbf{100.0}_{\textcolor{gray}{\pm\text{0.0}}}$
& $\underline{\text{56.7}}_{\textcolor{gray}{\pm\text{3.4}}}$ \\

\midrule
\multicolumn{14}{l}{\textit{2D Pretrain Policy}} \\
CLIP (\textit{ICML'21}) 
& $\text{95.7}_{\textcolor{gray}{\pm\text{1.2}}}$ 
& $\text{78.3}_{\textcolor{gray}{\pm\text{4.6}}}$ 
& $\text{88.7}_{\textcolor{gray}{\pm\text{4.0}}}$ 
& $\text{61.7}_{\textcolor{gray}{\pm\text{5.7}}}$ 
& $\underline{\text{92.3}}_{\textcolor{gray}{\pm\text{1.2}}}$ 
& $\text{72.3}_{\textcolor{gray}{\pm\text{5.7}}}$ 
& $\text{82.7}_{\textcolor{gray}{\pm\text{3.2}}}$ 
& $\text{70.3}_{\textcolor{gray}{\pm\text{3.1}}}$ 
& $\text{58.7}_{\textcolor{gray}{\pm\text{3.5}}}$ 
& $\text{42.0}_{\textcolor{gray}{\pm\text{0.0}}}$ 
& $\text{36.7}_{\textcolor{gray}{\pm\text{4.6}}}$ 
& $\text{75.0}_{\textcolor{gray}{\pm\text{3.4}}}$ 
& $\text{36.7}_{\textcolor{gray}{\pm\text{4.3}}}$ \\

R3M (\textit{CoRL'22}) 
& $\text{26.7}_{\textcolor{gray}{\pm\text{2.3}}}$ 
& $\text{9.3}_{\textcolor{gray}{\pm\text{2.3}}}$ 
& $\text{32.0}_{\textcolor{gray}{\pm\text{4.0}}}$ 
& $\text{12.0}_{\textcolor{gray}{\pm\text{0.0}}}$ 
& $\text{26.7}_{\textcolor{gray}{\pm\text{2.3}}}$ 
& $\text{22.7}_{\textcolor{gray}{\pm\text{2.3}}}$ 
& $\text{24.0}_{\textcolor{gray}{\pm\text{6.9}}}$ 
& $\text{21.0}_{\textcolor{gray}{\pm\text{2.6}}}$ 
& $\text{51.3}_{\textcolor{gray}{\pm\text{6.0}}}$ 
& $\underline{\text{76.7}}_{\textcolor{gray}{\pm\text{2.9}}}$ 
& $\text{70.0}_{\textcolor{gray}{\pm\text{2.1}}}$ 
& $\text{51.7}_{\textcolor{gray}{\pm\text{6.4}}}$ 
& $\text{56.3}_{\textcolor{gray}{\pm\text{4.1}}}$ \\

VC-1 (\textit{NeurIPS'23}) 
& $\text{91.7}_{\textcolor{gray}{\pm\text{3.5}}}$ 
& $\text{53.0}_{\textcolor{gray}{\pm\text{8.7}}}$ 
& $\text{73.3}_{\textcolor{gray}{\pm\text{3.8}}}$ 
& $\text{24.7}_{\textcolor{gray}{\pm\text{0.6}}}$ 
& $\text{25.7}_{\textcolor{gray}{\pm\text{2.1}}}$ 
& $\text{24.7}_{\textcolor{gray}{\pm\text{2.1}}}$ 
& $\text{70.3}_{\textcolor{gray}{\pm\text{9.6}}}$ 
& $\text{34.7}_{\textcolor{gray}{\pm\text{2.3}}}$ 
& $\text{10.0}_{\textcolor{gray}{\pm\text{8.7}}}$ 
& $\text{40.7}_{\textcolor{gray}{\pm\text{8.1}}}$ 
& $\text{58.3}_{\textcolor{gray}{\pm\text{4.4}}}$ 
& $\text{83.3}_{\textcolor{gray}{\pm\text{5.8}}}$ 
& $\text{38.3}_{\textcolor{gray}{\pm\text{3.3}}}$ \\

HRP (\textit{CoRL'24}) 
& $\text{96.3}_{\textcolor{gray}{\pm\text{2.9}}}$ 
& $\text{57.0}_{\textcolor{gray}{\pm\text{9.2}}}$ 
& $\text{63.0}_{\textcolor{gray}{\pm\text{7.2}}}$ 
& $\text{19.0}_{\textcolor{gray}{\pm\text{2.0}}}$ 
& $\text{37.7}_{\textcolor{gray}{\pm\text{2.3}}}$ 
& $\text{45.7}_{\textcolor{gray}{\pm\text{2.1}}}$ 
& $\text{62.3}_{\textcolor{gray}{\pm\text{12.7}}}$ 
& $\text{41.7}_{\textcolor{gray}{\pm\text{1.2}}}$ 
& $\text{62.0}_{\textcolor{gray}{\pm\text{5.0}}}$ 
& $\text{29.7}_{\textcolor{gray}{\pm\text{6.4}}}$ 
& $\text{65.0}_{\textcolor{gray}{\pm\text{4.8}}}$ 
& $\underline{\text{98.3}}_{\textcolor{gray}{\pm\text{2.6}}}$ 
& $\text{38.3}_{\textcolor{gray}{\pm\text{3.5}}}$ \\

MPI (\textit{RSS'24}) 
& $\underline{\text{98.0}}_{\textcolor{gray}{\pm\text{1.7}}}$ 
& $\text{40.7}_{\textcolor{gray}{\pm\text{2.9}}}$ 
& $\text{47.0}_{\textcolor{gray}{\pm\text{13.1}}}$ 
& $\text{17.3}_{\textcolor{gray}{\pm\text{2.6}}}$ 
& $\text{60.0}_{\textcolor{gray}{\pm\text{3.7}}}$ 
& $\underline{\text{81.7}}_{\textcolor{gray}{\pm\text{8.3}}}$ 
& $\text{68.0}_{\textcolor{gray}{\pm\text{9.0}}}$ 
& $\underline{\text{75.0}}_{\textcolor{gray}{\pm\text{1.7}}}$ 
& $\text{55.5}_{\textcolor{gray}{\pm\text{2.0}}}$ 
& $\text{71.7}_{\textcolor{gray}{\pm\text{1.4}}}$ 
& $\text{66.7}_{\textcolor{gray}{\pm\text{3.9}}}$ 
& $\text{53.7}_{\textcolor{gray}{\pm\text{5.8}}}$ 
& $\text{43.7}_{\textcolor{gray}{\pm\text{3.1}}}$ \\

MCR (\textit{ICLR'25}) 
& $\text{75.7}_{\textcolor{gray}{\pm\text{3.5}}}$ 
& $\text{50.3}_{\textcolor{gray}{\pm\text{9.6}}}$ 
& $\text{42.7}_{\textcolor{gray}{\pm\text{4.6}}}$ 
& $\text{20.0}_{\textcolor{gray}{\pm\text{0.0}}}$ 
& $\text{28.0}_{\textcolor{gray}{\pm\text{3.0}}}$ 
& $\text{26.3}_{\textcolor{gray}{\pm\text{2.1}}}$ 
& $\text{34.0}_{\textcolor{gray}{\pm\text{2.6}}}$ 
& $\text{24.7}_{\textcolor{gray}{\pm\text{4.0}}}$ 
& $\text{59.0}_{\textcolor{gray}{\pm\text{1.7}}}$ 
& $\text{71.3}_{\textcolor{gray}{\pm\text{4.0}}}$ 
& $\text{60.0}_{\textcolor{gray}{\pm\text{4.2}}}$ 
& $\text{46.7}_{\textcolor{gray}{\pm\text{5.1}}}$ 
& $\text{36.7}_{\textcolor{gray}{\pm\text{2.7}}}$ \\

R3M-A (\textit{CVPR'25}) 
& $\underline{\text{98.0}}_{\textcolor{gray}{\pm\text{0.0}}}$ 
& $\text{44.0}_{\textcolor{gray}{\pm\text{6.9}}}$ 
& $\text{32.3}_{\textcolor{gray}{\pm\text{4.5}}}$ 
& $\text{21.7}_{\textcolor{gray}{\pm\text{2.9}}}$ 
& $\text{31.0}_{\textcolor{gray}{\pm\text{5.3}}}$ 
& $\text{56.7}_{\textcolor{gray}{\pm\text{6.0}}}$ 
& $\text{50.3}_{\textcolor{gray}{\pm\text{2.1}}}$ 
& $\text{18.3}_{\textcolor{gray}{\pm\text{2.1}}}$ 
& $\underline{\text{63.3}}_{\textcolor{gray}{\pm\text{2.9}}}$ 
& $\text{72.7}_{\textcolor{gray}{\pm\text{0.6}}}$ 
& $\text{68.3}_{\textcolor{gray}{\pm\text{3.6}}}$ 
& $\textbf{100.0}_{\textcolor{gray}{\pm\text{0.0}}}$ 
& $\text{45.0}_{\textcolor{gray}{\pm\text{2.4}}}$ \\

DINOv3 (\textit{arXiv'25}) 
& $\textbf{100.0}_{\textcolor{gray}{\pm\text{0.0}}}$ 
& $\text{90.7}_{\textcolor{gray}{\pm\text{2.1}}}$ 
& $\underline{\text{96.3}}_{\textcolor{gray}{\pm\text{0.9}}}$ 
& $\underline{\text{75.7}}_{\textcolor{gray}{\pm\text{3.4}}}$ 
& $\text{80.0}_{\textcolor{gray}{\pm\text{5.0}}}$ 
& $\text{71.7}_{\textcolor{gray}{\pm\text{5.7}}}$ 
& $\underline{\text{85.0}}_{\textcolor{gray}{\pm\text{4.3}}}$ 
& $\text{60.3}_{\textcolor{gray}{\pm\text{5.9}}}$ 
& $\text{26.0}_{\textcolor{gray}{\pm\text{8.5}}}$ 
& $\text{33.3}_{\textcolor{gray}{\pm\text{10.3}}}$ 
& $\text{70.0}_{\textcolor{gray}{\pm\text{4.0}}}$ 
& $\textbf{100.0}_{\textcolor{gray}{\pm\text{0.0}}}$ 
& $\text{43.3}_{\textcolor{gray}{\pm\text{3.3}}}$ \\

VIP$^\ast$ (\textit{ICML'25}) 
& $\text{96.0}_{\textcolor{gray}{\pm\text{3.6}}}$ 
& $\underline{\text{91.0}}_{\textcolor{gray}{\pm\text{6.6}}}$ 
& $\text{63.2}_{\textcolor{gray}{\pm\text{5.0}}}$ 
& $\text{68.0}_{\textcolor{gray}{\pm\text{4.6}}}$ 
& $\text{56.0}_{\textcolor{gray}{\pm\text{1.0}}}$ 
& $\text{73.7}_{\textcolor{gray}{\pm\text{3.5}}}$ 
& $\text{79.7}_{\textcolor{gray}{\pm\text{3.5}}}$ 
& $\text{71.0}_{\textcolor{gray}{\pm\text{9.2}}}$ 
& $\text{30.0}_{\textcolor{gray}{\pm\text{1.7}}}$ 
& $\text{64.4}_{\textcolor{gray}{\pm\text{9.2}}}$ 
& $\text{73.7}_{\textcolor{gray}{\pm\text{3.2}}}$ 
& $\text{96.3}_{\textcolor{gray}{\pm\text{2.7}}}$ 
& $\text{55.0}_{\textcolor{gray}{\pm\text{2.6}}}$ \\

\midrule
\textbf{LSA}
& $\textbf{100.0}_{\textcolor{gray}{\pm\text{0.0}}}$ 
& $\textbf{93.3}_{\textcolor{gray}{\pm\text{2.3}}}$ 
& $\textbf{100.0}_{\textcolor{gray}{\pm\text{0.0}}}$ 
& $\textbf{85.3}_{\textcolor{gray}{\pm\text{2.3}}}$ 
& $\textbf{98.0}_{\textcolor{gray}{\pm\text{2.0}}}$ 
& $\textbf{84.0}_{\textcolor{gray}{\pm\text{0.0}}}$ 
& $\textbf{92.3}_{\textcolor{gray}{\pm\text{1.5}}}$ 
& $\textbf{82.7}_{\textcolor{gray}{\pm\text{1.5}}}$ 
& $\textbf{63.9}_{\textcolor{gray}{\pm\text{3.5}}}$ 
& $\textbf{78.3}_{\textcolor{gray}{\pm\text{1.6}}}$ 
& $\textbf{85.0}_{\textcolor{gray}{\pm\text{1.7}}}$ 
& $\textbf{100.0}_{\textcolor{gray}{\pm\text{0.0}}}$ 
& $\textbf{58.3}_{\textcolor{gray}{\pm\text{1.3}}}$ \\
\bottomrule
\end{tabular}
}
\vspace{-1.0em}
\end{table*}

As shown in Table~\ref{tab:sim_results}, LSA obtains consistently strong results across MetaWorld, DexArt, and Adroit.
On MetaWorld, LSA performs particularly well on challenging tasks such as \textit{pick-place-wall} and \textit{disassemble}, where accurate spatial perception is critical.
On DexArt and Adroit, LSA also achieves competitive or best performance on articulated-object and dexterous manipulation tasks.
Compared with direct 3D policy baselines such as DP3~\cite{dp} and KAN-We-Flow~\cite{kwf}, LSA attains these results without using explicit 3D observations during downstream policy learning, suggesting that the proposed pretraining effectively injects useful 3D-aware representations into the visual encoder.
We further evaluate robustness by reducing image brightness by 40\% and adding random colored circular occlusions during testing.
Tab.~\ref{tab:robustness} shows that LSA suffers smaller drops under these visual perturbations, decreasing by only 7.5\% under brightness reduction and 18.7\% under visual disturbance.
In contrast, the compared baselines show much larger degradation, indicating that LSA is more robust to superficial appearance changes.

\subsection{Ablation Results}
Ablation studies are conducted on four MetaWorld tasks, including \textit{dial turn}, \textit{hand insert}, \textit{pick place wall}, and \textit{disassemble}. We report the average evaluation success rate across these tasks.

\begin{table*}[t]
\centering
\scriptsize

\begin{minipage}[t]{0.47\textwidth}
\centering
\caption{
Robustness under visual perturbations.
}
\label{tab:robustness}
\vspace{1mm}
\renewcommand{\arraystretch}{1.12}
\setlength{\tabcolsep}{3.2pt}
\begin{tabularx}{\linewidth}{
@{}l
>{\centering\arraybackslash}X
>{\centering\arraybackslash}X
>{\centering\arraybackslash}X@{}}
\toprule
\textbf{Method} & \textbf{Perturb.} & \textbf{SR(\%)} & \textbf{Relative Drop} \\
\midrule
DINOv3~\cite{dinov3} 
& Bright. 
& 82.9 $\rightarrow$ 70.8 
& \textcolor{orange}{$\downarrow$14.6\%} \\
& Disturb. 
& 82.9 $\rightarrow$ 55.0 
& \textcolor{orange}{$\downarrow$33.7\%} \\
\midrule
R3M-A~\cite{miti}  
& Bright. 
& 36.8 $\rightarrow$ 20.6 
& \textcolor{orange}{$\downarrow$44.0\%} \\
& Disturb. 
& 36.8 $\rightarrow$ 18.7 
& \textcolor{orange}{$\downarrow$49.2\%} \\
\midrule
HRP~\cite{hrp}    
& Bright. 
& 44.0 $\rightarrow$ 13.8 
& \textcolor{orange}{$\downarrow$68.6\%} \\
& Disturb. 
& 44.0 $\rightarrow$ 31.0 
& \textcolor{orange}{$\downarrow$29.5\%} \\
\midrule
\textbf{LSA} 
& Bright. 
& \textbf{92.2 $\rightarrow$ 85.3} 
& \textcolor{orange}{$\downarrow$7.5\%} \\
& Disturb. 
& \textbf{92.2 $\rightarrow$ 75.0} 
& \textcolor{orange}{$\downarrow$18.7\%} \\
\bottomrule
\end{tabularx}
\end{minipage}
\hfill
\begin{minipage}[t]{0.49\textwidth}
\centering
\caption{
Ablation study of LSA's architectural components.
\textbf{MP}: motion planning; \textbf{RA}: representation alignment;
\textbf{SG}: structural guidance; \textbf{AG}: attentive guidance.
}
\label{tab:ablation}
\vspace{1mm}

{\tiny
\renewcommand{\arraystretch}{1.02}
\setlength{\tabcolsep}{0pt}
\begin{tabularx}{\linewidth}{
@{}
>{\centering\arraybackslash}X
|>{\centering\arraybackslash}X
>{\centering\arraybackslash}X
>{\centering\arraybackslash}X
>{\centering\arraybackslash}X
|>{\centering\arraybackslash}X
>{\centering\arraybackslash}X
@{}}
\toprule
 & \textbf{MP} & \textbf{RA} & \textbf{SG} & \textbf{AG}
 & \textbf{SR} & \textbf{Gain} \\
\midrule
Exp.1 & \ding{55} & \ding{55} & \ding{55} & \ding{55} & 73.3 & $+$0 \\
Exp.2 & \ding{51} & \ding{55} & \ding{55} & \ding{55} & 81.0 & $+$7.7 \\
Exp.3 & \ding{51} & \ding{51} & \ding{55} & \ding{55} & 85.2 & $+$11.9 \\
Exp.4 & \ding{51} & \ding{55} & \ding{51} & \ding{55} & 85.8 & $+$12.5 \\
Exp.5 & \ding{51} & \ding{55} & \ding{55} & \ding{51} & 86.3 & $+$13.0 \\
Exp.6 & \ding{51} & \ding{51} & \ding{51} & \ding{55} & 89.0 & $+$15.7 \\
Exp.7 & \ding{51} & \ding{55} & \ding{51} & \ding{51} & 88.4 & $+$15.1 \\
Exp.8 & \ding{51} & \ding{51} & \ding{51} & \ding{51}
& \textbf{92.2} & $+$\textbf{18.9} \\
\bottomrule
\end{tabularx}
}
\end{minipage}

\vspace{-1.2em}
\end{table*}

\textbf{Architecture Design.}
As shown in Table~\ref{tab:ablation}, the model without human-video pretraining achieves an average success rate of 73.3\%.
Adding the motion planning objective improves the success rate to 81.0\%, indicating that future 3D hand trajectories provide transferable action-relevant supervision.
Built upon this objective, each component in LSA further improves the learned representation: representation alignment, structural guidance, and attentive guidance individually improve the success rate to 85.2\%, 85.8\%, and 86.3\%, respectively.
Combining representation alignment with structural guidance further increases the success rate to 89.0\%, suggesting that depth-aware reconstruction further strengthens the dense 3D features learned from representation alignment.
Finally, the complete LSA framework achieves the best success rate of 92.2\%, improving the non-pretrained baseline by 18.9\%.
These results show that the three phases of LSA provide complementary supervision, empowering the policy with 3D-aware perception and interaction-aware motion planning abilities learned from human videos.

\begin{table*}[t]
    \centering
    \scriptsize
    \renewcommand{\arraystretch}{1.15}
    \setlength{\tabcolsep}{4pt}

    \begin{minipage}[t]{0.48\textwidth}
        \centering
        \caption{Ablation study on key LSA variants.}
        \label{tab:ablation_variants}
        \vspace{2mm}
        \begin{tabularx}{\linewidth}{@{}X c@{}}
            \toprule
            \textbf{Configuration} & \textbf{SR(\%)} \\
            \midrule
            \textbf{Full Model (Ours)} & \textbf{92.2} \\
            \midrule
            \rowcolor{lightgray}
            \multicolumn{2}{@{}l@{}}{\textit{Motion Supervision}} \\
            Only 3D Hand Motion & 81.0  \\
            Only 2D Hand Motion & 77.3  \\
            3D $\rightarrow$ 2D Hand Motion & 56.3 \textcolor{orange}{$(\downarrow 35.9)$} \\
            \midrule
            \rowcolor{lightgray}
            \multicolumn{2}{@{}l@{}}{\textit{Aligned Representation}} \\
            Only VGGT & 89.8 \textcolor{orange}{$(\downarrow 2.4)$} \\
            Only Depth Pro & 90.3 \textcolor{orange}{$(\downarrow 1.9)$} \\
            \midrule
            \rowcolor{lightgray}
            \multicolumn{2}{@{}l@{}}{\textit{Alignment Weight}} \\
            None & 89.3 \textcolor{orange}{$(\downarrow 2.9)$} \\
            Condition only on $\mathbf{c}_l$ & 91.2 \textcolor{orange}{$(\downarrow 1.0)$} \\
            Condition only on $\mathbf{c}_m$ & 90.1 \textcolor{orange}{$(\downarrow 2.1)$} \\
            \midrule
            \rowcolor{lightgray}
            \multicolumn{2}{@{}l@{}}{\textit{Attentive Guidance}} \\
            Target: Each head & 86.3 \textcolor{orange}{$(\downarrow 5.9)$} \\
            Super.: Random head & 83.5 \textcolor{orange}{$(\downarrow 8.7)$} \\
            Super.: Worst-matching head & 81.8 \textcolor{orange}{$(\downarrow 10.4)$} \\
            Super.: Hand heatmap & 84.3 \textcolor{orange}{$(\downarrow 7.9)$} \\
            \bottomrule
        \end{tabularx}
    \end{minipage}
    \hfill
    \begin{minipage}[t]{0.48\textwidth}
        \centering
        \caption{Quality analysis of 3D hand pseudo-labels. ADE denotes the average displacement error in meters.}
        \label{tab:pseudo_quality}
        \vspace{2mm}

        \textbf{(a) Wrist 3D lifting accuracy.}
        \vspace{0.25em}

        {
        \renewcommand{\arraystretch}{1.25}
        \setlength{\tabcolsep}{4pt}
        \begin{tabularx}{\linewidth}{@{}l c 
                                        >{\centering\arraybackslash}X 
                                        >{\centering\arraybackslash}X@{}}
            \toprule
            \textbf{Depth} & \textbf{Intrinsics} & \textbf{H2O ADE$\downarrow$} & \textbf{DexYCB ADE$\downarrow$} \\
            \midrule
            GT       & GT   & 0.0000 & 0.0000 \\
            GT       & VGGT & 0.0093 & 0.0079 \\
            \midrule
            Depth Pro & GT   & 0.0306 & 0.0333 \\
            Depth Pro & VGGT & 0.0393 & 0.0364 \\
            \midrule
            VGGT     & GT   & 0.0974 & 0.1147 \\
            VGGT     & VGGT & 0.0966 & 0.1155 \\
            \bottomrule
        \end{tabularx}
        }

        \vspace{1.0em}

        \textbf{(b) Effect of trajectory supervision quality.}
        \vspace{0.25em}

        {
        \renewcommand{\arraystretch}{1.25}
        \begin{tabularx}{\linewidth}{@{}X c@{}}
            \toprule
            \textbf{Trajectory Supervision} & \textbf{SR(\%)} \\
            \midrule
            VGGT-lifted trajectory & 88.6 \textcolor{orange}{$(\downarrow 3.6)$} \\
            Depth Pro + VGGT (Ours) & 92.2 \\
            DexYCB GT trajectory & 93.5 \textcolor{blue}{$(\uparrow 1.3)$} \\
            \bottomrule
        \end{tabularx}
        }
    \end{minipage}
    \vspace{-1em}
\end{table*}

\textbf{Analysis of LSA Variants.}
Tab.~\ref{tab:ablation_variants} analyzes key design choices in LSA.
For motion supervision, using only 2D hand motion achieves 77.3\%, slightly below the 3D motion planning baseline of 81.0\%.
However, when the model receives 3D geometric supervision in the Lift and See phases but uses 2D trajectory prediction in the Act phase, the performance drops further.
We attribute this to the semantic inconsistency between the lifted 3D representation space and the image-plane 2D motion objective.
Using only VGGT or Depth Pro for representation alignment is inferior to using both, indicating that different 3D FMs provide complementary dense geometric cues.
Removing dynamic alignment weights or conditioning them on a single layer also degrades performance, showing the need for adaptive multi-teacher weighting with multi-layer geometric context.
For attentive guidance, supervising the attention of all shallow heads suppresses attention diversity and thus degrades performance.
In addition, using a random head, the worst-matching head, or the hand heatmap as the supervision target also reduces performance, indicating that the internally emerged attention target selected by the hand heatmap more effectively captures interaction regions.

\textbf{Quality of 3D Pseudo-labels.}
Tab.~\ref{tab:pseudo_quality} evaluates the reliability of the pseudo-3D hand labels used by LSA.
To analyze the error, we separately evaluate lifting accuracy on H2O and DexYCB, using 240 valid wrist samples from 12 scenes for each dataset; more details are provided in \textbf{Appendix~\ref{app:lift_eval_set}}.
Depth Pro depth with VGGT intrinsics substantially reduces wrist lifting errors compared with VGGT depth variants, e.g., 0.0364\,m vs. 0.1155\,m on DexYCB.
When replacing only the Act-phase trajectory labels, noisier VGGT-lifted trajectories reduce SR from 92.2\% to 88.6\%, while DexYCB GT trajectories only improve SR to 93.5\%.
The limited improvement from replacing our pseudo-trajectories with GT trajectories suggests that the adopted Depth Pro+VGGT pipeline provides reasonably reliable supervision in our setting. We also validate the effectiveness of our labeling and pretraining strategy on more diverse unlabeled datasets, such as Ego4D (\textbf{Appendix~\ref{app:data_diversity}}).

\textbf{More Analyses.}
Given the limited space, we provide additional analyses in the \textbf{Appendix}, including detailed experimental protocols (\textbf{Appendix~\ref{app:exp_protocol}}), comparisons with direct 3D-FM features and additional pretrained baselines (\textbf{Appendix~\ref{app:direct_3dfm_comparison},~\ref{app:additional_baselines}}), teacher-selection studies (\textbf{Appendix~\ref{app:selection_3dfms}}), data scaling and diversity studies (\textbf{Appendix~\ref{app:data_scaling},~\ref{app:data_diversity}}), parameter update strategies (\textbf{Appendix~\ref{app:update_strategy}}), hyperparameter and layer-selection ablations (\textbf{Appendix~\ref{app:hyperparameter},~\ref{app:layer_selection}}), training convergence (\textbf{Appendix~\ref{app:loss_curves}}), computational overhead (\textbf{Appendix~\ref{app:computational_overhead}}), and more visualizations of attentive guidance and task trajectories (\textbf{Appendix~\ref{app:visualization}}).



\section{Real-World Experiments}
\label{sec:real}

\subsection{Setup}
\label{sec:real_setup}

\begin{figure}[t]
    \centering
    \includegraphics[width=\linewidth]{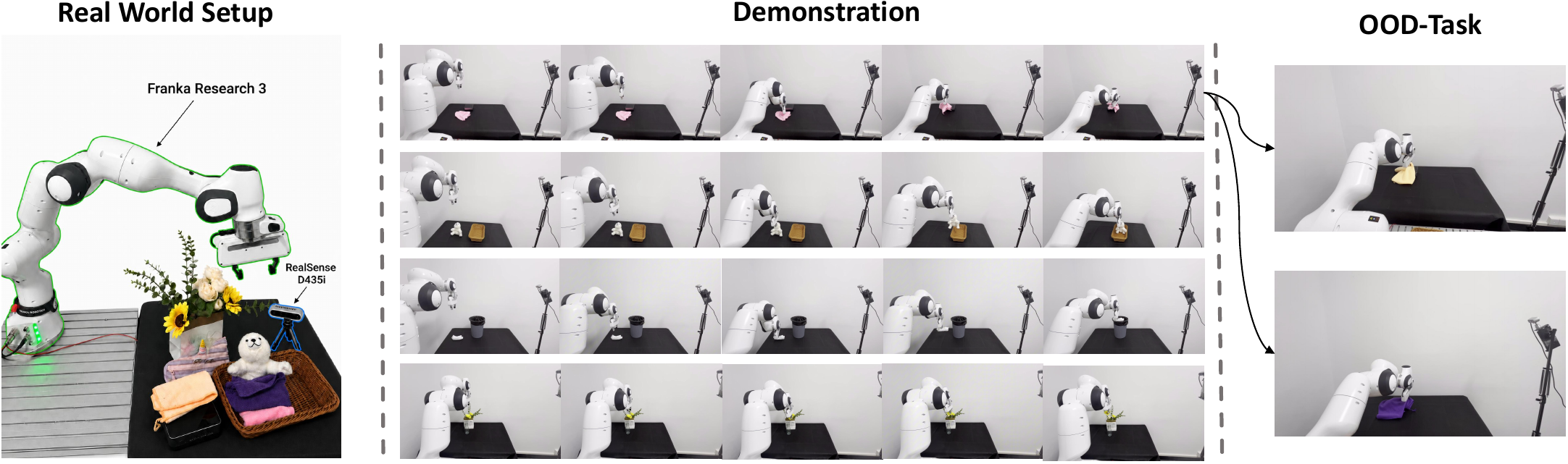}
    \caption{
    Overview of our real-world experiment setup.
    The figure shows our robotic platform and sensors (left), the execution of the four tasks under the in-distribution setup (middle), and the modifications applied for the out-of-distribution setup (right).
    }
    \label{fig:real_setup}
    \vspace{-1em}
\end{figure}

We further evaluate LSA in real-world manipulation using a 7-DoF Franka Research 3 robot equipped with a 1-DoF parallel gripper. 
RGB observations are captured by a RealSense D435i camera. 
As shown in Fig.~\ref{fig:real_setup}, we consider four daily-scenario manipulation tasks: \textit{Wipe PC}, \textit{Pick Trash}, \textit{Place Toy}, and \textit{Pluck Flower}. 
For each task, we collect 30 demonstrations with variations in the initial poses of the robot and objects. 
We evaluate each policy over ten trials per task and report the success rate (\%). 

To evaluate out-of-distribution (OOD) generalization, we further construct two unseen variants of \textit{Wipe PC}. 
The policy is trained with one towel color and tested with two previously unseen towel colors, denoted as \textit{unseen1} and \textit{unseen2}. 
This setting tests whether the learned representation can generalize beyond superficial appearance changes. 
More implementation details are provided in \textbf{Appendix~\ref{app:real_details}}.

\begin{table}[t]
\centering
\scriptsize
\caption{
Real-world evaluation with the Franka robot across four tasks.
For \textit{Wipe PC}, \textit{U1} and \textit{U2} denote two unseen towel colors.
}
\label{tab:real_world_results}
\resizebox{\linewidth}{!}{
\begin{tabular}{lccccccc}
\toprule
\textbf{Method} 
& \textbf{Pluck Flower} 
& \textbf{Pick Trash} 
& \textbf{Place Toy} 
& \textbf{Wipe PC} 
& \textbf{Wipe PC U1} 
& \textbf{Wipe PC U2} 
& \textbf{Avg.} \\
\midrule
DINOv3~\cite{dinov3} & 20.0 & 30.0 & 20.0 & 30.0 & 10.0 & 20.0 & 21.7 \\
R3M-A~\cite{miti}  & 10.0 & 30.0 & 30.0 & 40.0 & 0.0  & 10.0 & 20.0 \\
VIP~\cite{li2024vip}    & \textbf{50.0} & 40.0 & 50.0 & 40.0 & 20.0 & 20.0 & 36.7 \\
HRP~\cite{hrp}    & 40.0 & 40.0 & 60.0 & 50.0 & 20.0 & 40.0 & 41.7 \\
\rowcolor{lightgray}
\textbf{Ours} 
       & \textbf{50.0} 
       & \textbf{70.0} 
       & \textbf{80.0} 
       & \textbf{80.0} 
       & \textbf{70.0} 
       & \textbf{60.0} 
       & \textbf{68.3} \\
\bottomrule
\end{tabular}
}
\vspace{-2em}
\end{table}

\subsection{Evaluation Results}
\label{sec:real_results}

Tab.~\ref{tab:real_world_results} shows that LSA achieves the highest average success rate among all compared methods, with 68.3\% vs. 41.7\% for the strongest baseline HRP.
LSA outperforms the second-best method on \textit{Pick Trash} (70.0\% vs. 40.0\%), \textit{Place Toy} (80.0\% vs. 60.0\%), and \textit{Wipe PC} (80.0\% vs. 50.0\%), while matching the best result on \textit{Pluck Flower} (50.0\%).
On the two OOD \textit{Wipe PC} settings, LSA obtains 70.0\% and 60.0\%, whereas the best baselines achieve 20.0\% and 40.0\%; notably, R3M-A drops to 0.0\% on \textit{Wipe PC U1}.
These results suggest that LSA transfers better to real-world manipulation.
On one hand, its human-video pretraining learns 3D-aware representations from 2D observations, reducing the burden of acquiring spatial relations from limited robot demonstrations.
On the other hand, its interaction-focused guidance helps the policy preserve task-relevant hand-object relations under appearance changes.



\section{Conclusion}
\label{sec:conclusion}
In this paper, we propose LSA, a hierarchical robot policy pretraining framework that systematically decomposes the learning process into three synergistic phases: Lift, See, and Act. Unlike prior works that rely solely on sparse 3D trajectories, LSA fully exploits the multi-tiered knowledge of 3D FMs by aligning policy features with dense, intermediate geometric representations. To bridge the gap between implicit features and physical execution, we introduce a dual-guidance mechanism in the See phase, combining explicit structural reconstruction with attentive guidance to focus on task-relevant regions. Finally, the Act phase grounds the learned representation in future 3D hand motion, which serves as an action-relevant proxy for downstream robot manipulation. \textbf{Limitations.}
LSA shares several limitations inherent to human-video-based robot pretraining methods due to the human-robot embodiment gap.
\textbf{Broader Impacts}. Looking forward, this work opens avenues for scaling multi-tiered 3D knowledge extraction from human videos to broader robot manipulation scenarios, with the potential to further enhance Vision-Language-Action (VLA) model pretraining. To the best of our knowledge, this work has no obvious negative social impacts.

\section*{Acknowledgments}
This work was supported by the National Natural Science Foundation of China under Grant 62476099 and 62076101, Guangdong Basic and Applied Basic Research Foundation under Grant 2024B1515020082, the 2026 Guangdong Provincial Special Fund for Industrial Foundation Reengineering (Multimodal Adaptive Embodied AI Large Model Project), the Key-Area Research and Development Program of Guangdong Province under Grant 2025B0909020002 and 2024B0101040004, the TCL Young Scholars Program, and the High Performance Computing Platform of South China University of Technology.

{
\small
  \bibliographystyle{unsrtnat}  
  \bibliography{main}     
}


\appendix



\clearpage
\appendix

\noindent
\begin{minipage}{\linewidth}
\centering
{\Large \textbf{Appendix Contents}}
\vspace{0.8em}

\raggedright
\small

\noindent\hyperref[app:method_details]{\textbf{A. Additional Methodology Details}}
\dotfill \pageref{app:method_details}

\vspace{0.25em}
\noindent\hspace{1em}\hyperref[app:overall_loss]{A.1 Overall Pretraining Objective}
\dotfill \pageref{app:overall_loss}

\noindent\hspace{1em}\hyperref[app:motion_planning]{A.2 Motion Planning Loss}
\dotfill \pageref{app:motion_planning}

\noindent\hspace{1em}\hyperref[app:dynamic_weight]{A.3 Dynamic Weight Allocation}
\dotfill \pageref{app:dynamic_weight}

\noindent\hspace{1em}\hyperref[app:depth_reconstruction]{A.4 More Details of Depth Reconstruction}
\dotfill \pageref{app:depth_reconstruction}

\noindent\hspace{1em}\hyperref[app:att_guidance]{A.5 Attentive Guidance Details}
\dotfill \pageref{app:att_guidance}

\noindent\hspace{1em}\hyperref[app:data_cleaning]{A.6 Data Cleaning and 3D Lifting}
\dotfill \pageref{app:data_cleaning}

\noindent\hspace{1em}\hyperref[app:hand_prior]{A.7 Hand Prior Heatmap}
\dotfill \pageref{app:hand_prior}

\noindent\hspace{1em}\hyperref[app:pretrain_details]{A.8 Pretraining Implementation Details}
\dotfill \pageref{app:pretrain_details}

\vspace{0.6em}
\noindent\hyperref[app:exp_protocol]{\textbf{B. Experimental Protocols}}
\dotfill \pageref{app:exp_protocol}

\vspace{0.25em}
\noindent\hspace{1em}\hyperref[app:sim_setup]{B.1 Simulation Evaluation Setup}
\dotfill \pageref{app:sim_setup}

\noindent\hspace{1em}\hyperref[app:baseline_details]{B.2 Baseline Details}
\dotfill \pageref{app:baseline_details}

\noindent\hspace{1em}\hyperref[app:lift_eval_set]{B.3 3D Wrist Lifting Evaluation Set Construction}
\dotfill \pageref{app:lift_eval_set}

\noindent\hspace{1em}\hyperref[app:real_details]{B.4 Real-World Implementation Details}
\dotfill \pageref{app:real_details}

\vspace{0.6em}
\noindent\hyperref[app:additional_results]{\textbf{C. Additional Experimental Results}}
\dotfill \pageref{app:additional_results}

\vspace{0.25em}
\noindent\hspace{1em}\hyperref[app:direct_3dfm_comparison]{C.1 Comparison with Direct 3D-FM Features}
\dotfill \pageref{app:direct_3dfm_comparison}

\noindent\hspace{1em}\hyperref[app:selection_3dfms]{C.2 Selection of 3D FMs}
\dotfill \pageref{app:selection_3dfms}

\noindent\hspace{1em}\hyperref[app:data_scaling]{C.3 Data Scaling and Sample Efficiency}
\dotfill \pageref{app:data_scaling}

\noindent\hspace{1em}\hyperref[app:update_strategy]{C.4 Parameter Update Strategies}
\dotfill \pageref{app:update_strategy}

\noindent\hspace{1em}\hyperref[app:data_diversity]{C.5 Pretraining Data Diversity}
\dotfill \pageref{app:data_diversity}

\noindent\hspace{1em}\hyperref[app:hyperparameter]{C.6 Hyperparameter Selection}
\dotfill \pageref{app:hyperparameter}

\noindent\hspace{1em}\hyperref[app:layer_selection]{C.7 Layer Selection}
\dotfill \pageref{app:layer_selection}

\noindent\hspace{1em}\hyperref[app:loss_curves]{C.8 Training Loss Curves}
\dotfill \pageref{app:loss_curves}

\noindent\hspace{1em}\hyperref[app:additional_baselines]{C.9 Additional Baseline Comparisons}
\dotfill \pageref{app:additional_baselines}

\noindent\hspace{1em}\hyperref[app:real_world_generalization]{C.10 Additional Real-World Generalization Experiments}
\dotfill \pageref{app:real_world_generalization}

\noindent\hspace{1em}\hyperref[app:computational_overhead]{C.11 Computational Overhead}
\dotfill \pageref{app:computational_overhead}

\vspace{0.6em}
\noindent\hyperref[app:further_limitations]{\textbf{D. Further Discussion on Limitations}}
\dotfill \pageref{app:further_limitations}

\vspace{0.6em}
\noindent\hyperref[app:visualization]{\textbf{E. Additional Visualizations}}
\dotfill \pageref{app:visualization}

\vspace{0.25em}
\noindent\hspace{1em}\hyperref[app:sim_traj_vis]{E.1 Simulation Trajectory Visualizations}
\dotfill \pageref{app:sim_traj_vis}

\noindent\hspace{1em}\hyperref[app:real_traj_vis]{E.2 Real-World Trajectory Visualizations}
\dotfill \pageref{app:real_traj_vis}

\noindent\hspace{1em}\hyperref[app:more_ag_vis]{E.3 More Attentive Guidance Visualizations}
\dotfill \pageref{app:more_ag_vis}

\end{minipage}

\vspace{1em}

\section{Additional Methodology Details}
\label{app:method_details}

\subsection{Overall Pretraining Objective}
\label{app:overall_loss}

The overall supervision loss consists of four terms: the motion planning loss $\mathcal{L}_{\mathrm{motion}}$, the representation alignment loss $\mathcal{L}_{\mathrm{align}}$, the depth reconstruction loss $\mathcal{L}_{\mathrm{recon}}$, and the attentive guidance loss $\mathcal{L}_{\mathrm{att}}$.
The total loss in the pretraining stage is formulated as:
\begin{equation}
    \label{eq:overall_loss}
    \mathcal{L}_{\mathrm{total}} =
    \mathcal{L}_{\mathrm{motion}}
    +
    \lambda_{g}
    \cdot
    \left(
    \mathcal{L}_{\mathrm{align}}
    +
    \mathcal{L}_{\mathrm{recon}}
    +
    \lambda_{a}
    \cdot
    \mathcal{L}_{\mathrm{att}}
    \right),
\end{equation}
where $\lambda_{g}$ balances the geometry-related supervision terms and $\lambda_{a}$ controls the strength of attentive guidance.
In our experiments, we set $\lambda_{g}=0.3$ and $\lambda_{a}=0.5$.




\subsection{Motion Planning Loss}
\label{app:motion_planning}

We provide the explicit formulation of the motion planning objective and define all symbols below:
\begin{equation}
\mathcal{L}_{\mathrm{motion}}
=
\frac{1}{3L}\sum_{t=1}^{L}
\left\|\hat{\mathbf{T}}_t-\mathbf{T}_t\right\|_1
=
\frac{1}{3L}\sum_{t=1}^{L}
\left(
|\hat{x}_t-x_t|
+
|\hat{y}_t-y_t|
+
|\hat{z}_t-z_t|
\right),
\end{equation}
where $L$ is the number of predicted future timesteps; $t$ indexes these timesteps; $\hat{\mathbf{T}}_t=(\hat{x}_t,\hat{y}_t,\hat{z}_t)\in\mathbb{R}^{3}$ and $\mathbf{T}_t=(x_t,y_t,z_t)\in\mathbb{R}^{3}$ denote the predicted and lifted target 3D wrist positions at timestep $t$, respectively; and $\|\cdot\|_1$ denotes the sum of absolute errors over the three spatial coordinates. The objective then averages the absolute errors over the $L$ future timesteps and the three spatial coordinates.

\subsection{Dynamic Weight Allocation}
\label{app:dynamic_weight}
We design the dynamic weight assignment mechanism to efficiently harness representations from multiple 3D FMs, as the optimal contribution of each FM fundamentally varies across diverse manipulation scenarios.
To adaptively learn the most valuable knowledge, we formulate a condition query leveraging the \cls tokens $\mathbf{c}_l$ and $\mathbf{c}_m$ from the representation alignment layer ($l$-th layer) and the depth reconstruction layer ($m$-th layer), respectively.
As summarized in \textbf{Algorithm}~\ref{alg:weight_allocation}, this condition query is used to compute the scaled dot-product similarity with geometric tokens and produce the dynamic teacher weights.
First, the \cls token naturally serves as a compact global representation, effectively encapsulating the spatial semantics of the entire scene.
Second, the two selected \cls tokens capture geometry-related semantics.
Specifically, the $l$-th layer is directly aligned with the implicit intermediate representations of the 3D FMs, thus capturing \textbf{implicit geometric information}.
Conversely, the $m$-th layer is explicitly optimized for the depth reconstruction auxiliary task, thereby encoding the \textbf{explicit 3D spatial geometry} of the physical world.
By aggregating them, we formulate a condition query that achieves a comprehensive geometric awareness, thereby better modulating the teacher weights. Here, $\operatorname{MeanPool}$ denotes channel-group mean pooling: it reshapes the last feature dimension into groups of $D=512$ and averages over the group axis while preserving the batch and spatial-token dimensions.
\begin{algorithm}[htbp]
\centering
\begin{minipage}{\linewidth}
\caption{Dynamic Weight Allocation}
\label{alg:weight_allocation}
\small

    
    
    
\begin{algorithmic}[1]
    \renewcommand{\algorithmicrequire}{\textbf{Input:}}
    \renewcommand{\algorithmicensure}{\textbf{Output:}}
    \REQUIRE $\mathbf{c}_l, \mathbf{c}_m$ from $l$-th and $m$-th layers; geometric tokens $\mathbf{E}'_c, \mathbf{E}'_d$.
    \ENSURE Global modulation weights $w_c, w_d$.

    \STATE $\mathbf{Q} \leftarrow \mathrm{Unsqueeze}\!\left(\mathrm{MeanPool}(\mathrm{Concat}(\mathbf{c}_l, \mathbf{c}_m)), 1\right) \in \mathbb{R}^{B \times 1 \times D}$ \COMMENT{Query}
    
    \FORALL{$j \in \{c, d\}$}
        \STATE $\mathbf{\hat{E}}'_j \leftarrow \mathrm{MeanPool}(\mathbf{E}'_j) \in \mathbb{R}^{B \times N_a \times D}$ \COMMENT{Channel-group pooling}
        \STATE $\mathbf{S}_j \leftarrow \mathrm{Squeeze}\!\left((\mathbf{Q}\mathbf{\hat{E}}_j'^{\top})/\sqrt{D}, 1\right) \in \mathbb{R}^{B \times N_a}$
        \STATE $\mathbf{s}_j \leftarrow \frac{1}{N_a} \sum_{i=1}^{N_a} \mathbf{S}_{j,:,i} \in \mathbb{R}^{B}$ \COMMENT{Similarity}
    \ENDFOR
    
    \STATE $\mathbf{G} \leftarrow \mathrm{Softmax}([\mathbf{s}_c, \mathbf{s}_d]) \in \mathbb{R}^{B \times 2}$ \COMMENT{Gating}
    \STATE $w_c \leftarrow \mathrm{avg}(\mathbf{G}_{:,0})$; \quad $w_d \leftarrow \mathrm{avg}(\mathbf{G}_{:,1})$ \COMMENT{Aggregation}
    
    \RETURN $w_c, w_d$
\end{algorithmic}
\end{minipage}
\end{algorithm}

\subsection{More Details of Depth Reconstruction}
\label{app:depth_reconstruction}

The decoder takes the 196 patch tokens from encoder layer $m=6$, with shape $B\times196\times768$, corresponding to a $14\times14$ spatial grid. Each token is projected from 768 to 512 dimensions and augmented with a fixed 2D sine--cosine positional embedding. The resulting sequence is processed by eight self-attention Transformer blocks with 8 heads, preserving its shape of $B\times196\times512$. A linear head maps each token to 256 depth values representing one $16\times16$ patch, producing $B\times196\times256$. A non-overlapping Fold operation with kernel size and stride 16 assembles these patches into a $B\times1\times224\times224$ depth map. Thus, depth is predicted patchwise and supervised pixelwise over the full $224\times224$ resolution.

We resize the metric-depth target to $224\times224$, map its pixel values to $[0,1]$, and compute the mean squared error over all spatial locations:
\begin{equation}
\mathcal{L}_{\mathrm{recon}}
=
\frac{1}{BHW}
\sum_{b=1}^{B}
\sum_{p=1}^{HW}
\left(
\hat{D}_{(b,p)}-D_{g(b,p)}
\right)^2,
\end{equation}
where $B$ is the batch size; $H$ and $W$ are the height and width of the depth map; $b$ and $p$ index the samples and spatial pixels, respectively; $\hat{\mathbf{D}}$ denotes the predicted depth map; and $\mathbf{D}_g$ denotes the metric-depth target. The standard setting uses neither an invalid-pixel mask nor confidence filtering.

\subsection{Attentive Guidance Details}
\label{app:att_guidance}

\begin{figure*}[htbp]
    \centering
    \includegraphics[width=\linewidth]{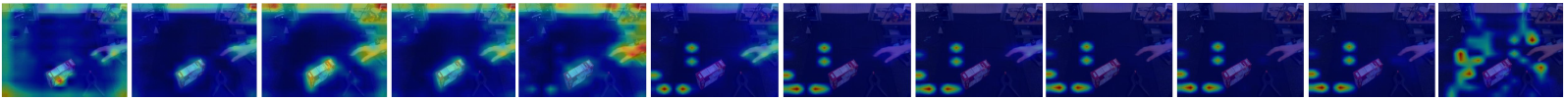}
    \caption{
    Attention maps of the \cls token from layers 1 to 12 in LSA without attentive guidance.
    Shallow layers tend to focus on local regions such as hands and objects, while deeper layers become more dispersed and abstract.
    }
    \label{fig:attention_layers_evolution}
    
\end{figure*}

\begin{figure}[t]
    \centering
    \includegraphics[width=0.6\linewidth]{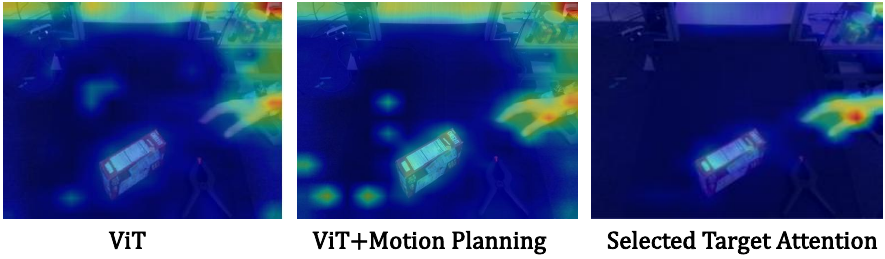}
    \caption{
Comparison of attention maps.
\textbf{Left:} averaged attention map over the first $R=6$ layers of the original ViT.
\textbf{Middle:} averaged attention map after motion-planning pretraining.
\textbf{Right:} target attention map selected by the hand prior.
The selected target attends to both the hand and the interacting object.
}
    \label{fig:compare}
    
\end{figure}

We further provide details and visual evidence for attentive guidance.
We treat the \cls tokens from different ViT layers as distinct entities, since they encode hierarchical global semantics.
As shown in Fig.~\ref{fig:attention_layers_evolution}, the attention maps associated with these layer-specific \cls tokens also evolve across layers.
Shallow layers tend to focus on local regions such as hands and objects, whereas deeper layers become more abstract and dispersed.
Motivated by this layer-wise pattern, we apply attentive guidance to the shallow layers. 
As shown in Fig.~\ref{fig:compare}, the original ViT produces relatively dispersed shallow attention, while motion planning encourages clearer attention to the hand region and further induces emergent attention to the interacting object.
This suggests that future hand trajectories provide supervision signals that implicitly capture hand-object spatial relations.
We therefore use the hand prior to select the internal attention map most aligned with the hand region; as shown in Fig.~\ref{fig:compare}, the selected target also extends to the interacting object.
Compared with supervising each individual head, our method guides only the head-averaged attention map of each shallow layer, aligning the overall spatial focus while preserving inter-head diversity.




\subsection{Data Cleaning and 3D Lifting}
\label{app:data_cleaning}

We use DexYCB as the main pretraining data source.
Our data cleaning process consists of three steps.

First, we detect the hand pose in each frame.
Due to occlusions and detection failures, we remove a video if the hand is not detected for $\geq 3$ consecutive frames.

Second, given a per-frame metric depth sequence $\{z_t\}_{t=1}^{T}$ in a video, we compute its temporal standard deviation
$\sigma_z=\mathrm{std}(\{z_t\})$
and frame-to-frame depth differences
$\Delta z_t=|z_t-z_{t-1}|$ for $t=2,\ldots,T$.
A video is discarded as low quality if it is globally unstable,
i.e.,
$\sigma_z>0.09$ and
$\frac{1}{T-1}\sum_{t=2}^{T}\Delta z_t>0.02$,
or if it contains an abrupt jump,
i.e.,
$\max_t \Delta z_t>0.15$.
These thresholds are set empirically to remove unstable lifted trajectories.

Third, we apply Gaussian smoothing to the metric depth sequence $\{z_t\}$ with \(\sigma_{\mathrm t}=3.0\) and obtain the smoothed depth sequence $\{\tilde{z}_t\}$.
The first and last depth values are kept unchanged to avoid boundary artifacts.

After data cleaning, we lift each 2D wrist keypoint $(u_t,v_t)$ into 3D space using the smoothed metric depth $\tilde{z}_t$ and camera intrinsics.
Let $(f_x,f_y)$ denote the focal lengths and $(c_x,c_y)$ denote the principal point.
The lifted 3D wrist point $\mathbf{p}_t$ in the camera coordinate system is computed by the pinhole camera model:
\begin{equation}
\mathbf{p}_t=
\begin{bmatrix}
X_t\\
Y_t\\
Z_t
\end{bmatrix}
=
\begin{bmatrix}
(u_t-c_x)\tilde{z}_t/f_x\\
(v_t-c_y)\tilde{z}_t/f_y\\
\tilde{z}_t
\end{bmatrix},
\quad t=1,\dots,T.
\end{equation}

\subsection{Hand Prior Heatmap}
\label{app:hand_prior}

\begin{figure}[t]
  \centering
  \includegraphics[width=0.5\linewidth]{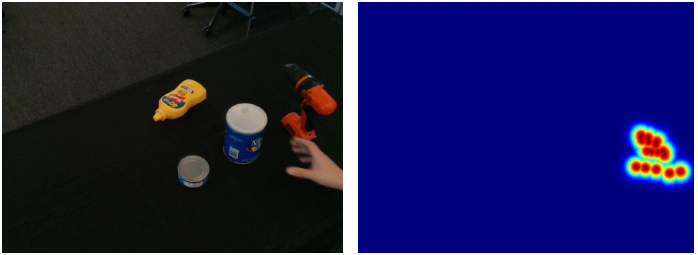}
  \caption{
  \textbf{Left:} Original image.
  \textbf{Right:} Generated hand prior heatmap.
  }
  \label{fig:hand_prior}
\end{figure}

To construct the spatial hand prior used in attentive guidance, we employ MediaPipe~\cite{mediapipe} to extract the 2D image coordinates of 21 hand joints.
Let $\mathbf{p}_k=(x_k,y_k)$ denote the 2D coordinate of the $k$-th hand joint.
We place a Gaussian kernel at each detected keypoint and aggregate the responses using the maximum value at each pixel location:
\begin{equation}
H(x,y)=
\max_{k\in\{1,\dots,21\}}
\exp\left(
-\frac{(x-x_k)^2+(y-y_k)^2}{2\sigma_{\mathrm h}^2}
\right).
\end{equation}
Here, $\sigma_{\mathrm h}$ controls the spatial spread of the local response.
We further convert $H$ into a softened spatial prior using a temperature-scaled Softmax:
\begin{equation}
h(x,y)=
\frac{\exp(H(x,y)/\tau_{\mathrm h})}
{\sum_{i,j}\exp(H(i,j)/\tau_{\mathrm h})}.
\end{equation}
Here, $\tau_{\mathrm h}$ denotes the Softmax temperature.
This operation smooths the distribution and allows the hand prior to cover not only exact keypoint locations but also nearby hand regions and interaction context. 
We then min--max normalize $h$ to $[0,1]$ and resize it to the $14\times14$ patch grid before computing the attention-alignment scores.
In our experiments, we set $\sigma_{\mathrm h}=12$ and $\tau_{\mathrm h}=5$. Fig.~\ref{fig:hand_prior} shows the results.

\subsection{Pretraining Implementation Details}
\label{app:pretrain_details}

We adopt a pretrained ViT-B~\cite{vit} initialized on ImageNet~\cite{imagenet} as the visual encoder.
To enable parameter-efficient adaptation, we keep the original ViT-B parameters frozen and apply LoRA~\cite{lora} with rank $r=8$ and scaling factor $\alpha=16$.
Following our hierarchical design, we perform representation alignment at the $l=3$ layer of the visual encoder, while the visual tokens from the $m=6$ layer are routed to an 8-layer Transformer decoder for metric depth reconstruction.
Attentive guidance is applied to the first $R=6$ layers. Attentive guidance is enabled after a 10-epoch warm-up with the motion planning objective, so that the model can first develop initial task-relevant attention patterns before the target attention map is selected.
For the Act phase, the model predicts the future 3D wrist trajectory for the next $L=3$ frames.

Following prior work~\cite{hoi}, we use DexYCB~\cite{dexycb} to pretrain LSA.
After data cleaning, we retain 20K video clips for pretraining.
Each clip consists of consecutive frames that depict human hands performing tasks and moving within the scene.
Optimization is performed using Adam with a cosine learning rate schedule and an initial learning rate of $5\times 10^{-4}$.
The model is trained for 50 epochs on NVIDIA RTX 4090 GPUs with 24GB memory each, using a batch size of 16 per GPU.

\section{Experimental Protocols}
\label{app:exp_protocol}

This section provides detailed experimental protocols for simulation benchmarks, baseline implementations, 3D wrist lifting evaluation, and real-world experiments.

\subsection{Simulation Evaluation Setup}
\label{app:sim_setup}

We provide detailed simulation settings for the three benchmarks used in the main paper: MetaWorld~\cite{meta}, DexArt~\cite{dexart}, and Adroit~\cite{adroit}.

\textbf{MetaWorld.}
MetaWorld provides tabletop manipulation tasks with a Sawyer arm and gripper.
We select eight tasks spanning different difficulty levels~\cite{maskworld}: regular tasks including \textit{hammer} and \textit{dial turn}, hard tasks including \textit{assembly} and \textit{hand insert}, and very hard tasks including \textit{pick place wall}, \textit{shelf place}, \textit{disassemble}, and \textit{stick pull}.
For each task, we use scripted policies to collect 25 expert demonstrations.
Following prior works~\cite{175-1,175-2}, the maximum rollout length is strictly limited to 175 steps, which makes the setting more challenging by requiring faster and more accurate task completion.
The behavior cloning policy is trained for 20,000 steps with a batch size of 32 and evaluated every 2,000 steps.

\textbf{DexArt.}
DexArt focuses on dexterous manipulation of articulated objects using an XArm6 arm and an Allegro hand.
We evaluate on two tasks: \textit{bucket} and \textit{toilet}.
For each task, we collect 100 expert demonstrations using pretrained reinforcement learning policies.
Each rollout is capped at 100 steps.
The behavior cloning policy is trained for 3,000 epochs with a batch size of 64 and evaluated every 200 epochs.
The pretrained visual encoder is frozen during downstream policy learning, and only the policy head is trained.

\textbf{Adroit.}
Adroit simulates a multi-fingered Shadow Hand in MuJoCo~\cite{mujoco} and contains contact-rich dexterous manipulation tasks involving rigid and articulated objects.
We evaluate on three tasks: \textit{door}, \textit{hammer}, and \textit{pen}.
For each task, we collect 10 expert demonstrations using a pretrained reinforcement learning policy~\cite{vrl3}.
Each demonstration contains 100 steps.
The training and evaluation protocols follow the DexArt setting.
We include this benchmark to further evaluate whether LSA can transfer to complex dexterous manipulation with fine-grained hand-object interactions.

\textbf{Reporting Protocol.}
For each evaluation checkpoint, we compute the success rate over all rollouts.
For each random seed, we average the success rates of the top-3 performing checkpoints.
We then report the mean and sample standard deviation over three random seeds, rounded to one decimal place.

\subsection{Baseline Details}
\label{app:baseline_details}

We compare LSA with three categories of baselines.

\textbf{General-purpose Visual Representations.}
We include CLIP~\cite{clip} and DINOv3~\cite{dinov3} as strong general visual representation baselines.
CLIP is a vision-language model trained with contrastive image-text supervision, which learns transferable semantic visual features from large-scale image-caption pairs.
DINOv3 is a self-supervised vision foundation model that learns strong image-level and dense visual features without manual labels.
Although both models provide powerful general-purpose representations, their pretraining objectives are not explicitly designed for action-relevant 3D reasoning or robot manipulation.

\textbf{Robot Visual Pretraining Methods.}
We compare with MCR~\cite{mcr}, VIP~\cite{li2024vip}, R3M~\cite{r3m}, VC-1~\cite{vc1}, HRP~\cite{hrp}, MPI~\cite{mpi}, and R3M-A~\cite{miti}.
These methods represent strong robot-oriented visual pretraining baselines, covering different ways of learning action-relevant representations for downstream robotic control.

\textbf{Direct 3D Policies.}
We further compare with DP3~\cite{dp} and KAN-We-Flow~\cite{kwf}, both of which directly take 3D point clouds as policy inputs and learn action prediction through diffusion- or flow-matching-based objectives.
These methods provide direct comparisons to LSA from the perspective of 3D-aware policy learning. Following Lift3D~\cite{lift3d}, we compare the 2D and 3D policies under the same evaluation protocol, using $224\times224$ RGB images and single-view point clouds with 1,024 points as inputs, respectively, while retaining the original method-specific settings of the 3D policies.

For all 2D baselines, we use the same downstream demonstrations. The pretrained visual encoder is frozen and combined with the same lightweight behavior cloning policy head.
The policy head is implemented as a 3-layer MLP with hidden size 256, following prior works~\cite{mcr,hrp}. We apply the reporting protocol used for LSA to all 2D baselines.

\subsection{3D Wrist Lifting Evaluation Set Construction}
\label{app:lift_eval_set}

To quantify the noise of the lifted 3D wrist trajectories, we construct two small evaluation sets from H2O~\cite{h2o} and DexYCB~\cite{dexycb}, respectively.
For each dataset, we randomly sample 240 valid wrist samples from 12 scenes, covering either left or right hands.
Both datasets provide calibrated cameras and ground-truth 3D hand poses.
For each sample, we use the ground-truth 2D wrist projection as the wrist location, estimate depth and camera intrinsics from different sources, and lift the wrist point to the camera coordinate system.
Following~\cite{ade}, we report the 3D Average Displacement Error (ADE) between the lifted wrist and the ground-truth 3D wrist.

\subsection{Real-World Implementation Details}
\label{app:real_details}

\textbf{Hardware Setup.}
We conduct real-world experiments using a 7-DoF Franka Research 3 robot equipped with a 1-DoF parallel gripper.
A RealSense D435i camera is used to capture RGB observations from a fixed third-person viewpoint.
During data collection, we record robot states and RGB observations at 16 Hz.

\textbf{Demonstration Collection.}
For each task, we collect 30 teleoperated demonstrations.
The demonstrations include variations in the initial poses of the robot and objects to avoid learning a fixed execution trajectory.
The four real-world tasks are defined as follows:
\begin{itemize}
    \item \textit{Wipe PC}: the robot grasps a pink towel and wipes the surface of a laptop.
    \item \textit{Pick Trash}: the robot picks up a piece of trash and drops it into a trash bin.
    \item \textit{Place Toy}: the robot grasps a toy and places it into a basket.
    \item \textit{Pluck Flower}: the robot grasps a flower and lifts it upward.
\end{itemize}

\textbf{Policy Training.}
Following prior real-world policy evaluation protocols, we freeze the pretrained visual encoder and train only the downstream policy head.
The policy head is a 3-layer MLP with two hidden layers of size 512.
We train the policy using a log-likelihood objective with a learning rate of $1\times10^{-4}$ and a batch size of 128 for 50k steps.
The predicted actions are linearly interpolated to 1000 Hz to satisfy the Franka control frequency.

\textbf{Evaluation Protocol.}
For each task, we evaluate each policy over 10 independent trials and report the success rate.
A trial is considered successful if the robot completes the target manipulation goal without human intervention.
For the OOD variants of \textit{Wipe PC}, the policy is trained with a pink towel and evaluated with two unseen towel colors, purple and yellow.
This setting changes the object appearance while preserving the same task goal and interaction structure.

\section{Additional Experimental Results}
\label{app:additional_results}

This section provides additional experimental analyses that are not included in the main paper due to space limitations.
Unless otherwise specified, ablation studies are conducted on four MetaWorld tasks, including \textit{dial turn}, \textit{hand insert}, \textit{pick place wall}, and \textit{disassemble}. 
We report the average evaluation success rate across these tasks.

\subsection{Comparison with Direct 3D-FM Features}
\label{app:direct_3dfm_comparison}

We compare Full LSA with variants that directly use 3D-FM features. We construct direct Depth Pro, VGGT, and dual-FM policies by uniformly projecting their frozen features to 768 dimensions and feeding them into the same action head. We also combine these direct features with LSA without Lift. All variants use the same downstream training and evaluation settings as Full LSA. The results are shown in Table~\ref{tab:direct_3dfm_comparison}.

\begin{table}[H]
\centering
\caption{Comparison between Full LSA and the direct use of 3D-FM features.}
\label{tab:direct_3dfm_comparison}
\footnotesize
\renewcommand{\arraystretch}{1.15}
\setlength{\tabcolsep}{2pt}
\begin{tabularx}{\linewidth}{
@{}
>{\raggedright\arraybackslash}X
ccccc
@{}}
\toprule
\textbf{Method} &
\textbf{SR (\%) $\uparrow$} &
\textbf{Params (M) $\downarrow$} &
\textbf{GFLOPs $\downarrow$} &
\textbf{FPS $\uparrow$} &
\textbf{Slowdown $\downarrow$} \\
\midrule
\textbf{Full LSA}
& \textbf{92.2}
& \textbf{86.1}
& \textbf{35.1}
& \textbf{230.6}
& \textbf{1.0$\times$} \\
Direct Depth Pro
& 40.7
& 953.0
& 382.3
& 59.6
& 3.9$\times$ \\
Direct VGGT
& 29.8
& 911.0
& 493.6
& 18.7
& 12.3$\times$ \\
Direct Dual-FM
& 26.9
& 1,863.7
& 875.9
& 14.5
& 15.9$\times$ \\
LSA w/o Lift + Depth Pro
& 81.4
& 1,039.4
& 417.5
& 47.0
& 4.9$\times$ \\
LSA w/o Lift + VGGT
& 61.6
& 997.3
& 528.8
& 17.5
& 13.2$\times$ \\
LSA w/o Lift + Dual-FM
& 66.4
& 1,950.1
& 911.1
& 13.6
& 17.0$\times$ \\
\bottomrule
\end{tabularx}
\end{table}

Full LSA achieves 92.2\%, whereas the three direct policies achieve only 40.7\%, 29.8\%, and 26.9\%. Adding direct features to LSA without Lift improves the results but remains below Full LSA, with the strongest variant reaching 81.4\%. Full LSA also deploys only 86.1M parameters and runs at 230.6 FPS, while the direct variants require 911.0M--1,950.1M parameters and are 3.9$\times$--17.0$\times$ slower. These results show that directly deploying generic 3D-FM features provides neither effective action representations nor practical inference efficiency. Lift instead transfers their dense 3D knowledge into a compact visual encoder during pretraining, yielding substantially stronger performance without running either 3D FM. All inference costs are measured on a single RTX 4090 with batch size 1 and one $224\times224$ RGB frame. Slowdown is measured relative to Full LSA.




\subsection{Selection of 3D FMs}
\label{app:selection_3dfms}

{
\setlength{\columnsep}{6pt}
\setlength{\intextsep}{2pt}

\begin{wraptable}{r}{0.43\columnwidth}
\vspace{-6pt}
\centering
\caption{Teacher combinations in the Lift phase.}
\label{tab:teacher_selection}
\vspace{-4pt}
\footnotesize
\renewcommand{\arraystretch}{1.05}
\begin{tabular*}{0.95\linewidth}{@{\extracolsep{\fill}}lc@{}}
\toprule
\textbf{Teacher Combination} & \textbf{SR (\%)} \\
\midrule
VGGT + Depth Pro  & 92.2 \\
DA3 + Depth Pro   & 92.0 \\
VGGT + UniDepthV2 & \textbf{92.5} \\
DA3 + UniDepthV2  & 91.8 \\
\bottomrule
\end{tabular*}
\vspace{-6pt}
\end{wraptable}

We use multiple 3D FMs to obtain complementary 3D supervision. VGGT provides camera parameters and geometric representations, while Depth Pro provides absolute-depth information. Their combination provides richer and complementary representations than either teacher alone. This design is not restricted to specific 3D FMs. We further replace VGGT and Depth Pro individually or jointly with Depth Anything 3 (DA3)~\cite{da3} and UniDepthV2~\cite{unidepth}. As shown in Table~\ref{tab:teacher_selection}, replacing the teachers individually or jointly achieves comparable performance, demonstrating that LSA is compatible with alternative 3D FMs.

}

\subsection{Data Scaling and Sample Efficiency}
\label{app:data_scaling}

\begin{figure}[t]
    \centering
    \includegraphics[width=0.8\linewidth]{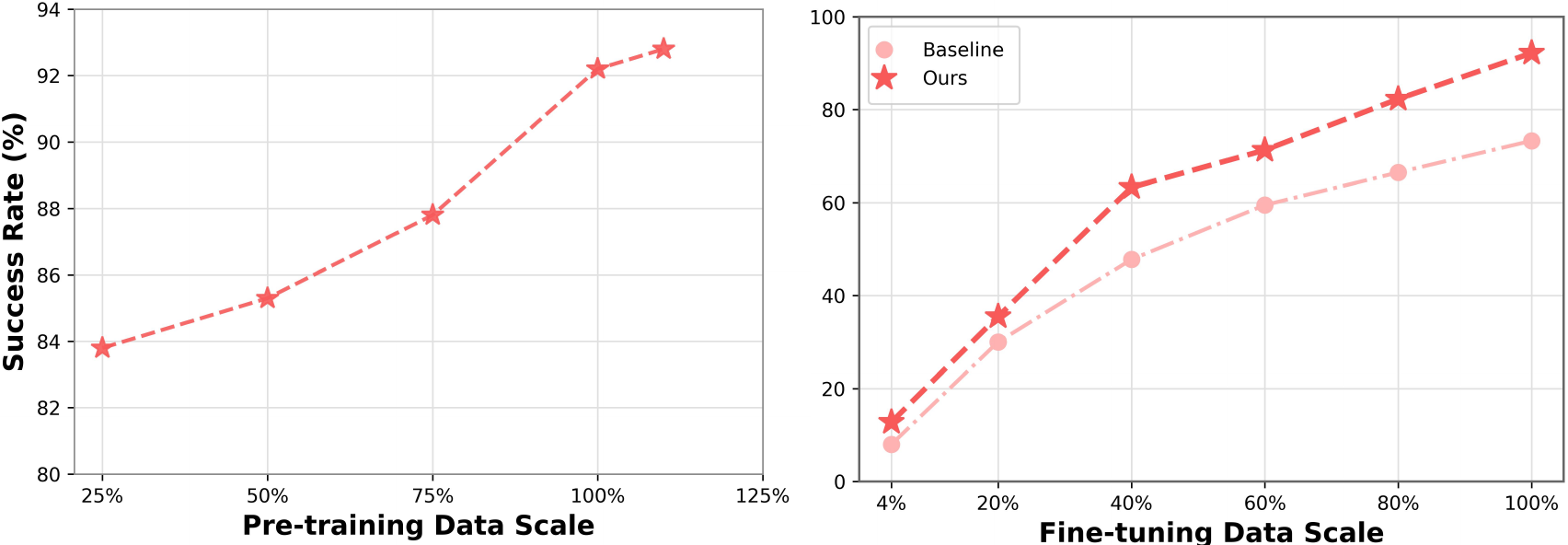}
    
    \caption{Influence of pretraining and fine-tuning data scale.}
    \label{fig:data_eff}
   
\end{figure}

We first investigate the effect of data scale during both human-video pretraining and downstream policy learning.
As shown in Fig.~\ref{fig:data_eff}, increasing the amount of pretraining video data consistently improves downstream success rates, indicating that LSA can benefit from larger-scale human-video data.
We also evaluate the sample efficiency of downstream policy learning by varying the number of expert demonstrations.
LSA nearly matches the peak performance of baseline methods using only 60\% of the downstream expert demonstrations.
This suggests that the multi-tiered 3D knowledge learned during pretraining reduces the burden of learning spatial and interaction cues from limited robot demonstrations.

\subsection{Parameter Update Strategies}
\label{app:update_strategy}

We compare different parameter update strategies during pretraining.
As shown in Tab.~\ref{tab:update_strategy}, updating all parameters leads to a large performance drop.
This may be caused by overfitting to the relatively limited pretraining data and forgetting the general visual priors of the initialized ViT.
Fine-tuning only LayerNorm parameters performs much better, but still underperforms LoRA.
Therefore, we adopt LoRA as the default parameter-efficient adaptation strategy, as it adapts the encoder to our 3D supervision while preserving the general visual representation ability of the pretrained backbone.



\subsection{Pretraining Data Diversity}
\label{app:data_diversity}

We further evaluate whether LSA benefits from more diverse human-video sources.
As shown in Tab.~\ref{tab:data_source}, all settings use the same number of sampled clips, while the number of source videos varies.
Pretraining on H2O~\cite{h2o} with around 100 source videos achieves 90.2\%, while DexYCB with around 400 source videos improves the success rate to 92.2\%.
Using Ego4D \cite{ego4d} with around 1000 source videos further improves the result to 93.1\%.
These results suggest that LSA benefits from broader source-video diversity, indicating its potential to scale with more diverse human-video pretraining data.


\begin{table*}[t]
\centering

\begin{minipage}[t]{0.47\textwidth}
\centering
\caption{Comparison of different parameter update strategies.}
\label{tab:update_strategy}
\small
\renewcommand{\arraystretch}{1.15}
\setlength{\tabcolsep}{6pt}
\begin{tabular}{lc}
\toprule
\textbf{Method} & \textbf{SR (\%)} \\
\midrule
Full Parameters
& 67.2 \textcolor{orange}{$(\downarrow 25.0)$} \\
LN Fine-tuning
& 90.1 \textcolor{orange}{$(\downarrow 2.1)$} \\
LoRA Fine-tuning
& \textbf{92.2} \\
\bottomrule
\end{tabular}
\end{minipage}
\hfill
\begin{minipage}[t]{0.47\textwidth}
\centering
\caption{Effect of pretraining data diversity. All settings use 20K sampled clips.}
\label{tab:data_source}
\small
\renewcommand{\arraystretch}{1.15}
\setlength{\tabcolsep}{6pt}
\begin{tabular}{lcc}
\toprule
\textbf{Data Source} & \textbf{\#Videos} & \textbf{SR (\%)} \\
\midrule
H2O    & $\sim$100  & 90.2 \\
DexYCB & $\sim$400  & 92.2 \\
Ego4D  & $\sim$1000 & \textbf{93.1} \\
\bottomrule
\end{tabular}
\end{minipage}

\end{table*}

\subsection{Hyperparameter Selection}
\label{app:hyperparameter}

\begin{figure}[t]
    \centering
    \includegraphics[width=0.9\linewidth]{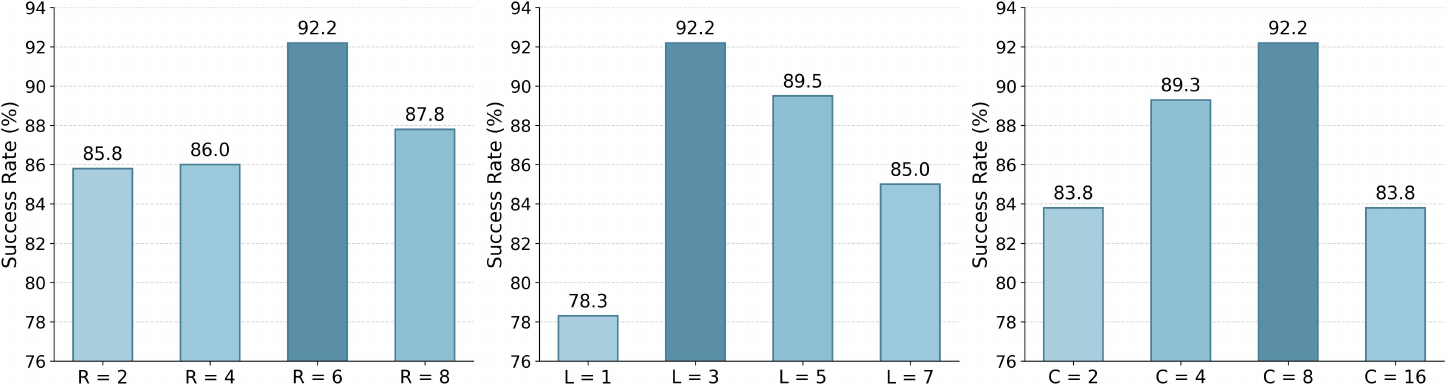}
    
    \caption{Impact of attentive guidance depth $R$, motion planning steps $L$, and the number of depth reconstruction decoder layers $C$.}
    \label{fig:hyper}
   
\end{figure}

As shown in Fig.~\ref{fig:hyper}, we investigate the optimal configuration of three critical hyperparameters: the depth of attentive guidance $R$, the number of motion planning steps $L$, and the number of depth reconstruction decoder layers $C$. 
For attentive guidance, applying it to the first $R=6$ layers achieves the highest success rate of $92.2\%$. Restricting guidance to only the first two layers ($R=2$) lowers the success rate to $85.8\%$, because this limited supervision makes it difficult to identify a reliable target attention map and leaves other optimizable shallow layers unguided. In contrast, extending it too deeply ($R=8$) reduces performance to $87.8\%$, as it limits the ability of deeper layers to learn abstract task semantics.
For motion planning, $L=3$ yields the best performance. A shorter horizon ($L=1$) severely degrades the success rate to $78.3\%$, because such a short horizon lacks sufficient temporal context to represent meaningful hand motion. In contrast, a longer horizon ($L=7$) reduces performance to $85.0\%$, since predicting farther-future trajectories becomes overly difficult and inaccurate. 
Finally, an 8-layer depth decoder ($C=8$) best balances model capacity and training stability. A shallower decoder ($C=2$) leads to a lower success rate of $83.8\%$, because its limited capacity is insufficient to reconstruct dense and accurate depth maps. In contrast, an overly deep decoder ($C=16$) also reduces performance to $83.8\%$, as it introduces optimization difficulties that hinder overall training. 
Therefore, we adopt $R=6$, $L=3$, and $C=8$ as our default setup.

\subsection{Layer Selection}
\label{app:layer_selection}

Tab.~\ref{tab:layer_ablation} presents an ablation on the layer configurations for representation alignment and depth reconstruction.
While keeping the depth reconstruction fixed at the 6th layer, setting the alignment layer too shallow—such as the (1, 6) configuration—leads to a 2.7\% drop in success rate, as the network has not yet extracted sufficient semantic features to effectively align with the high-dimensional implicit representations of the 3D FMs. Conversely, setting the alignment layer too deep, as seen in the (5, 6) setup, causes a 3.6\% performance decrease by excessively compressing the subsequent feature space required for explicit structural perception.

Similarly, fixing the feature alignment at the 3rd layer and altering the depth reconstruction layer also significantly impacts performance. Introducing reconstruction too early, as in the (3, 4) configuration, results in a 3.9\% decline in success rate. At this stage, the network has not fully assimilated the implicit 3D representations from the Lift phase, preventing it from effectively comprehending complex spatial layouts. Furthermore, delaying the reconstruction task to an excessively deep layer like (3, 8) triggers a severe performance drop of 7.6\%. This is because representations become highly abstract as they approach the Act phase, focusing primarily on global motion trajectory planning. Forcing these deep layers to retain fine-grained, pixel-level geometric information inherently interferes with the action planning task.

In addition, reversing the order to (6, 3) results in a 2.9\% drop. The model performs better when it first acquires a high-level 3D understanding (implicit) before attempting detailed geometric reconstruction (explicit). This suggests that the aligned 3D representations provide a necessary foundation for learning accurate 3D spatial layouts.
Finally, forcing the model to perform both feature alignment and depth reconstruction at the exact same layer—such as in the (3, 3) or (6, 6) configurations—results in substantial performance drops of 6.4\% and 9.9\%, respectively. Since feature alignment and depth reconstruction are tasks of completely different granularities, forcing them to share the same feature space triggers severe feature interference.

\begin{table*}[t]
\centering
\caption{Ablation study on the selection of alignment and reconstruction layers.}
\label{tab:layer_ablation}
\renewcommand{\arraystretch}{1.15}
\setlength{\tabcolsep}{4pt}

\resizebox{0.9\textwidth}{!}{
\begin{tabular}{lcccccccc}
\toprule
\multicolumn{1}{c}{
\multirow{2}{*}{
\shortstack{\textbf{Layer}\\\textbf{(Align, Recon)}}
}
}
&
\multicolumn{1}{c}{\textbf{Ours}} &
\multicolumn{2}{c}{\textbf{Varying Alignment}} &
\multicolumn{2}{c}{\textbf{Varying Reconstruction}} &
\multicolumn{1}{c}{\textbf{Reversed Order}} &
\multicolumn{2}{c}{\textbf{Shared Layer}} \\
\cmidrule(lr){2-2}
\cmidrule(lr){3-4}
\cmidrule(lr){5-6}
\cmidrule(lr){7-7}
\cmidrule(lr){8-9}

&
\textbf{(3, 6)}
& (1, 6)
& (5, 6)
& (3, 4)
& (3, 8)
& (6, 3)
& (3, 3)
& (6, 6) \\
\midrule

\textbf{SR (\%)}
& \textbf{92.2}
& 89.5 \textcolor{orange}{$(\downarrow 2.7)$}
& 88.6 \textcolor{orange}{$(\downarrow 3.6)$}
& 88.3 \textcolor{orange}{$(\downarrow 3.9)$}
& 84.6 \textcolor{orange}{$(\downarrow 7.6)$}
& 89.3 \textcolor{orange}{$(\downarrow 2.9)$}
& 85.8 \textcolor{orange}{$(\downarrow 6.4)$}
& 82.3 \textcolor{orange}{$(\downarrow 9.9)$} \\
\bottomrule
\end{tabular}
}
\end{table*}

\subsection{Training Loss Curves}
\label{app:loss_curves}

\begin{figure}[t]
  \centering
  \begin{subfigure}[b]{0.24\textwidth}
    \centering
    \includegraphics[width=\textwidth]{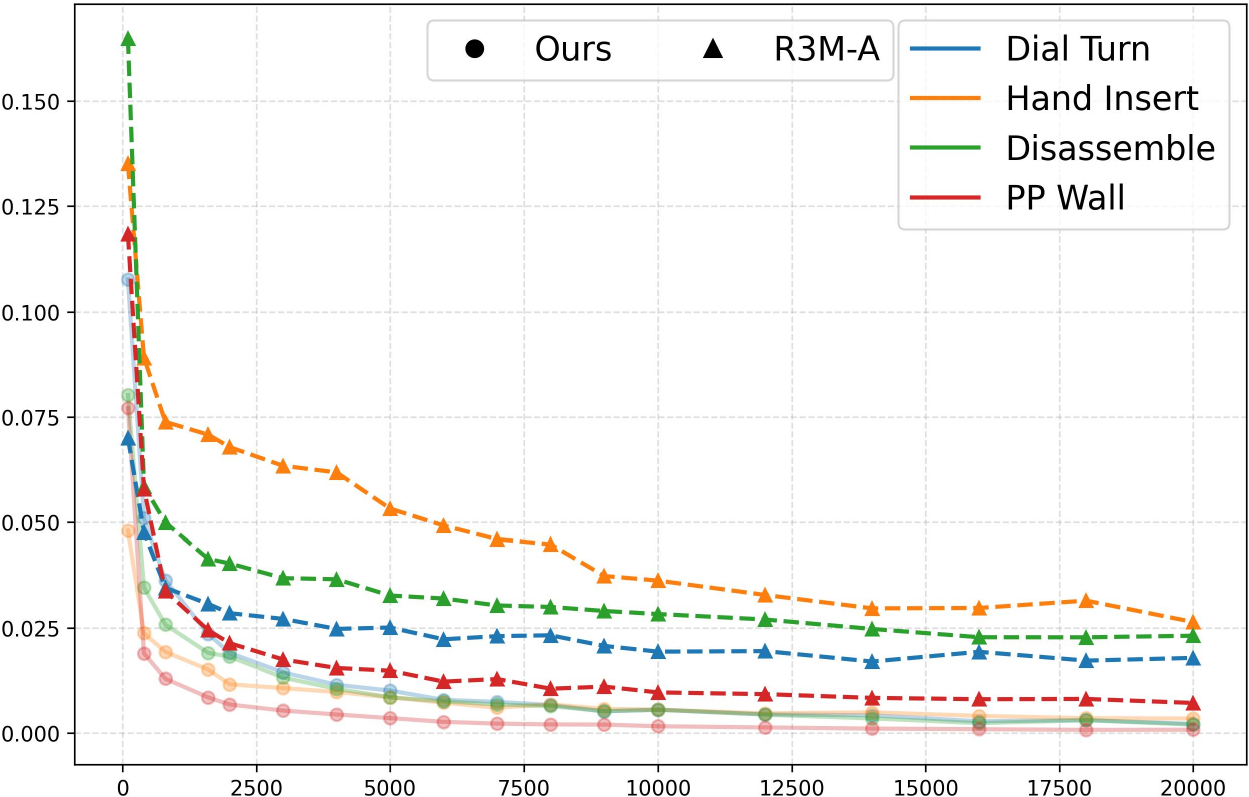}
    \caption{vs.\ R3M-A}
    \label{fig:vs_r3ma}
  \end{subfigure}
  \hfill
  \begin{subfigure}[b]{0.24\textwidth}
    \centering
    \includegraphics[width=\textwidth]{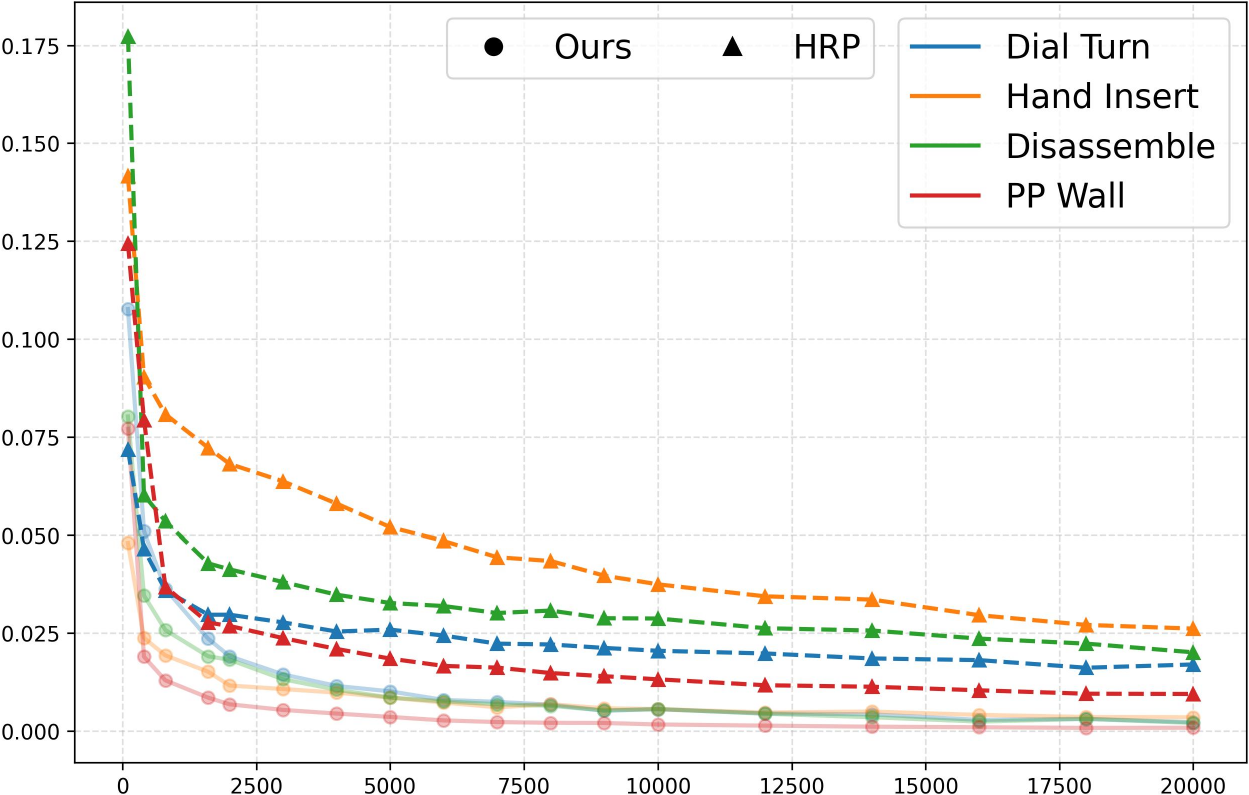}
    \caption{vs.\ HRP}
    \label{fig:vs_hrp}
  \end{subfigure}
  \hfill
  \begin{subfigure}[b]{0.24\textwidth}
    \centering
    \includegraphics[width=\textwidth]{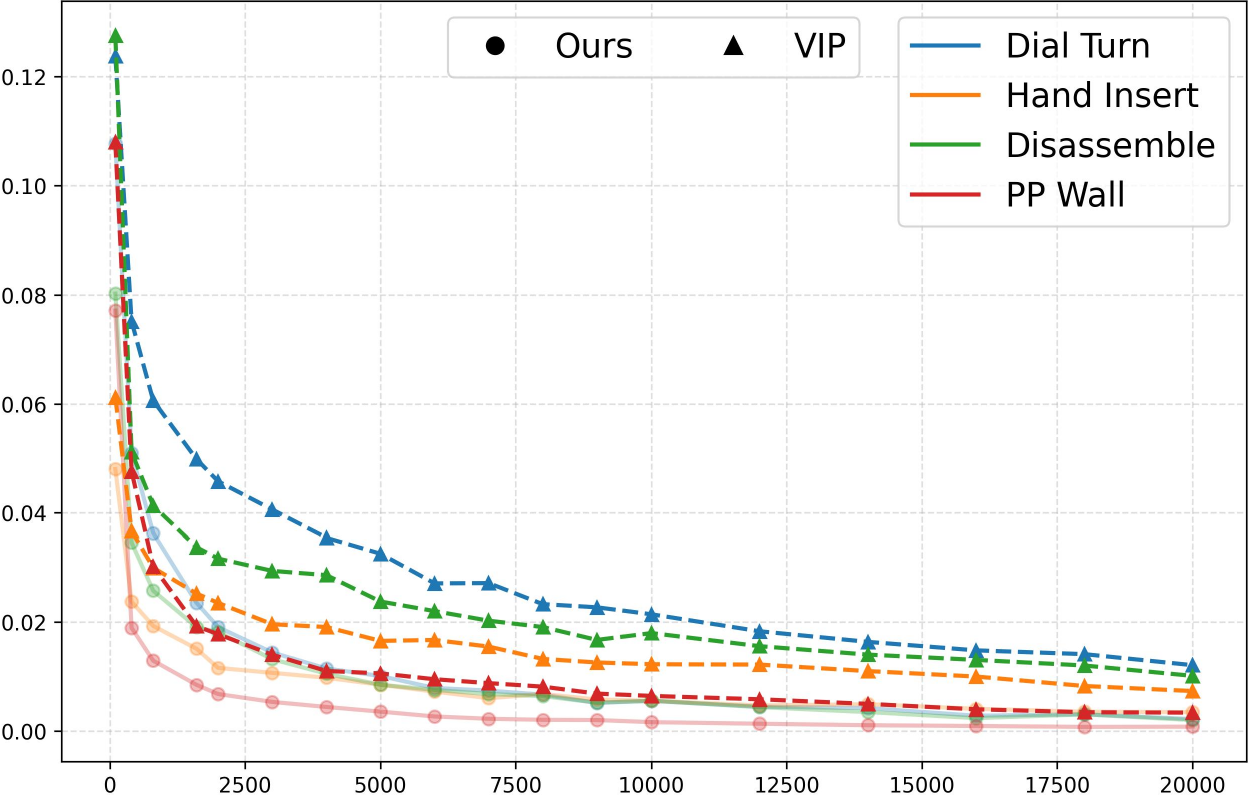}
    \caption{vs.\ VIP}
    \label{fig:vs_vip}
  \end{subfigure}
  \hfill
  \begin{subfigure}[b]{0.24\textwidth}
    \centering
    \includegraphics[width=\textwidth]{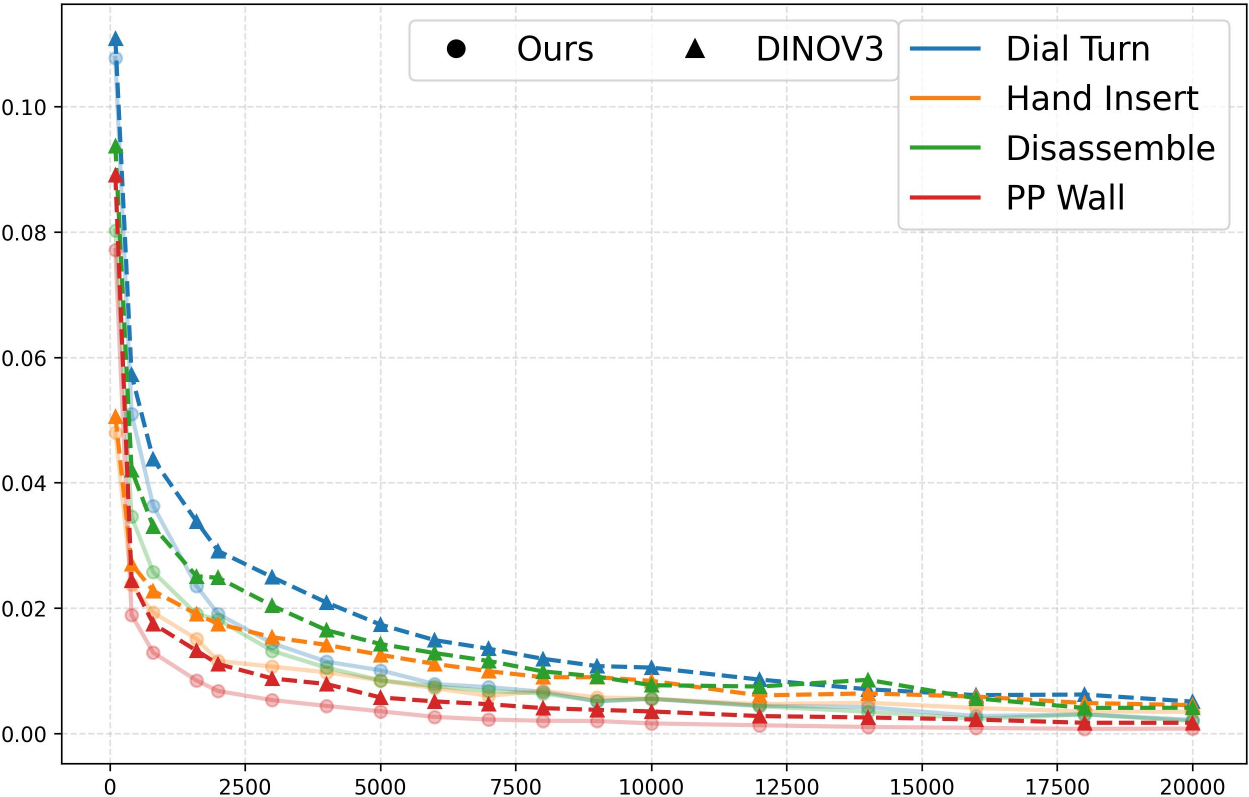}
    \caption{vs.\ DINOv3}
    \label{fig:vs_dinov3}
  \end{subfigure}
  \caption{Comparison of training convergence between LSA and several baselines: (a) R3M-A, (b) HRP, (c) VIP, and (d) DINOv3.}
  \label{fig:loss}
  
\end{figure}

Fig.~\ref{fig:loss} presents the fine-tuning loss curves of different pretraining methods on MetaWorld. Our method exhibits the fastest and most stable convergence and reaches the lowest loss among all compared approaches. We attribute this advantage to the multi-tiered 3D supervision introduced during pretraining, which equips the policy with more generalizable spatial and motion-planning knowledge and thus enables faster adaptation to downstream manipulation tasks.

\subsection{Additional Baseline Comparisons}
\label{app:additional_baselines}

We further compare LSA with the representative 3D-FM-based approach VITRA~\citep{vitra} and recent robotic pretraining methods, including H-RDT~\cite{hrdt} and Moto~\cite{motolatent}. To fairly evaluate their pretrained representations, we freeze the pretrained components of each method and use the same action head, downstream demonstrations, and evaluation protocol. As shown in Table~\ref{tab:additional_baselines_metaworld}, LSA achieves the highest average performance on MetaWorld among these methods.

\begin{table*}[t]
\centering
\caption{Additional baseline comparisons on MetaWorld.}
\label{tab:additional_baselines_metaworld}
\scriptsize
\setlength{\tabcolsep}{3.5pt}
\begin{tabular*}{\textwidth}{@{\extracolsep{\fill}}lccccccccc@{}}
\toprule
\textbf{Method}
& \textbf{HA}
& \textbf{DT}
& \textbf{AS}
& \textbf{HI}
& \textbf{PPW}
& \textbf{ShP}
& \textbf{DI}
& \textbf{StP}
& \textbf{Avg.} \\
\midrule
VITRA~\cite{vitra}
& $87.1{\pm}2.0$
& $82.2{\pm}6.7$
& $98.7{\pm}1.4$
& $77.8{\pm}2.0$
& $81.8{\pm}6.8$
& $77.8{\pm}2.0$
& $77.3{\pm}2.3$
& $78.2{\pm}2.8$
& 82.6 \\
Moto~\cite{motolatent}
& $21.8{\pm}3.3$
& $64.0{\pm}2.3$
& $87.6{\pm}3.4$
& $58.2{\pm}0.8$
& $57.8{\pm}4.7$
& $64.0{\pm}2.3$
& $64.4{\pm}4.3$
& $80.5{\pm}2.0$
& 62.3 \\
H-RDT~\cite{hrdt}
& $98.2{\pm}1.6$
& $86.2{\pm}2.0$
& $100.0{\pm}0.0$
& $83.5{\pm}2.0$
& $86.2{\pm}2.0$
& $76.0{\pm}2.3$
& $78.2{\pm}1.6$
& $92.4{\pm}4.1$
& 87.6 \\
\textbf{LSA (Ours)}
& $\mathbf{100.0{\pm}0.0}$
& $\mathbf{93.3{\pm}2.3}$
& $\mathbf{100.0{\pm}0.0}$
& $\mathbf{85.3{\pm}2.3}$
& $\mathbf{98.0{\pm}2.0}$
& $\mathbf{84.0{\pm}0.0}$
& $\mathbf{92.3{\pm}1.5}$
& $\mathbf{82.7{\pm}1.5}$
& \textbf{92.0} \\
\bottomrule
\end{tabular*}
\end{table*}

\subsection{Additional Real-World Generalization Experiments}
\label{app:real_world_generalization}

We extend the real-world OOD evaluation to Wipe PC and Place Toy. Background Change uses an unseen tablecloth, while Low-Light uses reduced ambient illumination unseen during training. Each policy is trained using the same 30 demonstrations, and each condition is evaluated over 20 independent trials. We compare LSA with HRP under identical settings. The results are reported in Table~\ref{tab:real_world_generalization}. LSA consistently achieves higher OOD success rates and smaller performance drops than HRP, extending the evaluation beyond object-appearance changes to unseen tabletop backgrounds and illumination conditions.

\begin{table*}[t]
\centering
\caption{Real-world generalization under unseen backgrounds and low-light conditions. Values in parentheses denote absolute performance drops.}
\label{tab:real_world_generalization}
\footnotesize
\renewcommand{\arraystretch}{1.15}
\setlength{\tabcolsep}{3pt}
\begin{tabular*}{\textwidth}{@{\extracolsep{\fill}}lcccccccc@{}}
\toprule
&
\multicolumn{3}{c}{\textbf{Place Toy}} &
\multicolumn{3}{c}{\textbf{Wipe PC}} &
\multicolumn{2}{c}{\textbf{Average Drop}} \\
\cmidrule(lr){2-4}
\cmidrule(lr){5-7}
\cmidrule(lr){8-9}
\textbf{Method}
& \textbf{Original}
& \textbf{Background}
& \textbf{Low-Light}
& \textbf{Original}
& \textbf{Background}
& \textbf{Low-Light}
& \textbf{Background}
& \textbf{Low-Light} \\
\midrule
HRP
& 65.0
& 35.0 ($-30.0$)
& 30.0 ($-35.0$)
& 60.0
& 35.0 ($-25.0$)
& 25.0 ($-35.0$)
& $-27.5$
& $-35.0$ \\
\textbf{LSA}
& \textbf{80.0}
& \textbf{65.0 ($-15.0$)}
& \textbf{60.0 ($-20.0$)}
& \textbf{75.0}
& \textbf{60.0 ($-15.0$)}
& \textbf{50.0 ($-25.0$)}
& \textbf{$-15.0$}
& \textbf{$-22.5$} \\
\bottomrule
\end{tabular*}
\end{table*}

\subsection{Computational Overhead}
\label{app:computational_overhead}

All computational costs are measured using NVIDIA RTX 4090 GPUs. The one-time offline processing of approximately 20K training sequences with VGGT and Depth Pro requires 0.56 GPU-hours and 36.37 GiB aggregate peak GPU memory, producing a 28.34 GiB feature cache. The 3D FMs are not used during downstream inference. LSA requires 4.34 ms per inference step with 0.341 GiB peak GPU memory.

\section{Further Discussion on Limitations}
\label{app:further_limitations}

\paragraph{Embodiment Differences.}
LSA adopts future 3D wrist trajectories as action proxies primarily to bypass the embodiment gap between humans and robots. In our preliminary experiments, additionally supervising with fine-grained 3D hand meshes, such as MANO~\cite{mano}, degraded performance compared with using wrist trajectories alone. We attribute this to the structural mismatch between human fingers and robot end-effectors, such as parallel grippers. Embodiment-specific action representations impose an additional burden on the downstream policy rather than providing transferable guidance. A promising solution is to extract latent actions from human hand motion, so that the supervision captures high-level motion intent rather than embodiment-specific action forms.

\paragraph{Object Interaction Discrepancies.}
Although LSA's attentive guidance effectively transfers region-level attention from human hand--object interactions to task-relevant gripper--object regions, it primarily provides coarse spatial guidance. Extending this supervision to finer-grained interaction patterns is more challenging because differences in embodiment appearance and manipulation modes may limit the direct transfer of such detailed cues to robots. One potential solution is to incorporate explicit object-affordance supervision, such as object regions and candidate contact points. Such object-centric and embodiment-agnostic signals could complement the current attentive guidance and further strengthen the transfer of interaction knowledge to downstream robot tasks.

\paragraph{Real-World Distribution Shifts.}
While our results suggest that 3D representations learned from human videos benefit downstream tasks, scene discrepancies between human-video data and real-world robot environments may still limit their transferability. A possible solution is to jointly pretrain on unlabeled human and real-robot videos using shared depth-reconstruction and 3D-feature-alignment objectives, encouraging the visual encoder to learn a unified 3D scene-geometric representation across both domains.

\section{Additional Visualizations}
\label{app:visualization}

This section provides additional visualizations.
We include representative simulation and real-world trajectories, as well as more attentive guidance visualizations on both human-object interaction and robotic manipulation scenes.
\subsection{Simulation Trajectory Visualizations}
\label{app:sim_traj_vis}

\begin{figure}[htbp]
    \centering
    \includegraphics[width=\linewidth]{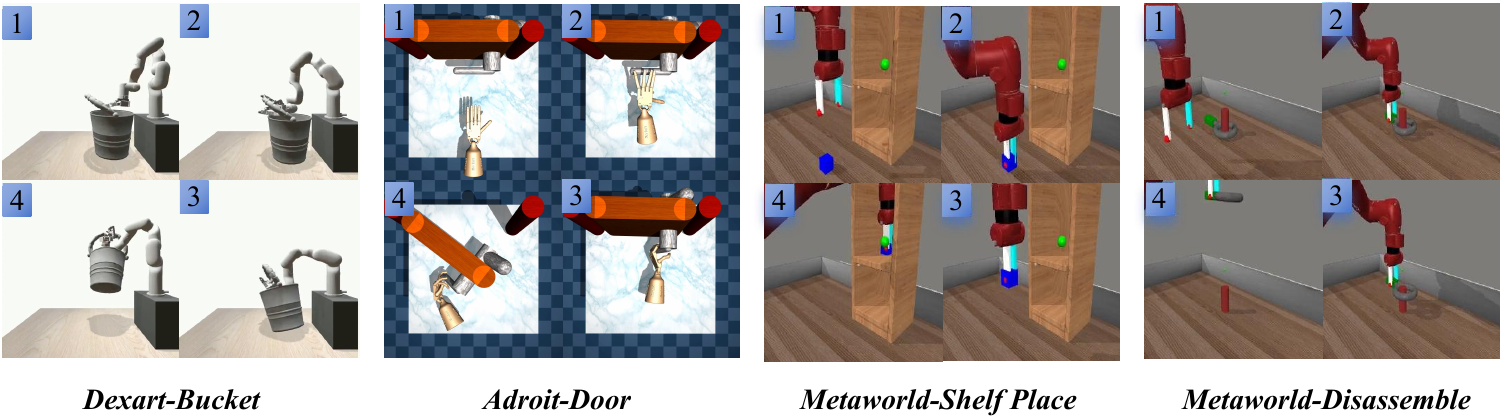}
    
    \caption{
    Visualization of representative simulation task trajectories.
    These examples illustrate the execution behavior of LSA in simulated manipulation environments.
    }
    \label{fig:sim}
   
\end{figure}

Fig.~\ref{fig:sim} shows representative task trajectories in simulation.
These examples qualitatively demonstrate that the policy pretrained by LSA can execute diverse manipulation behaviors across different simulated environments.
The visualized trajectories cover both tabletop manipulation and dexterous manipulation settings, complementing the quantitative results reported in the main paper.

\subsection{Real-World Trajectory Visualizations}
\label{app:real_traj_vis}

\begin{figure}[htbp]
    \centering
    \includegraphics[width=\linewidth]{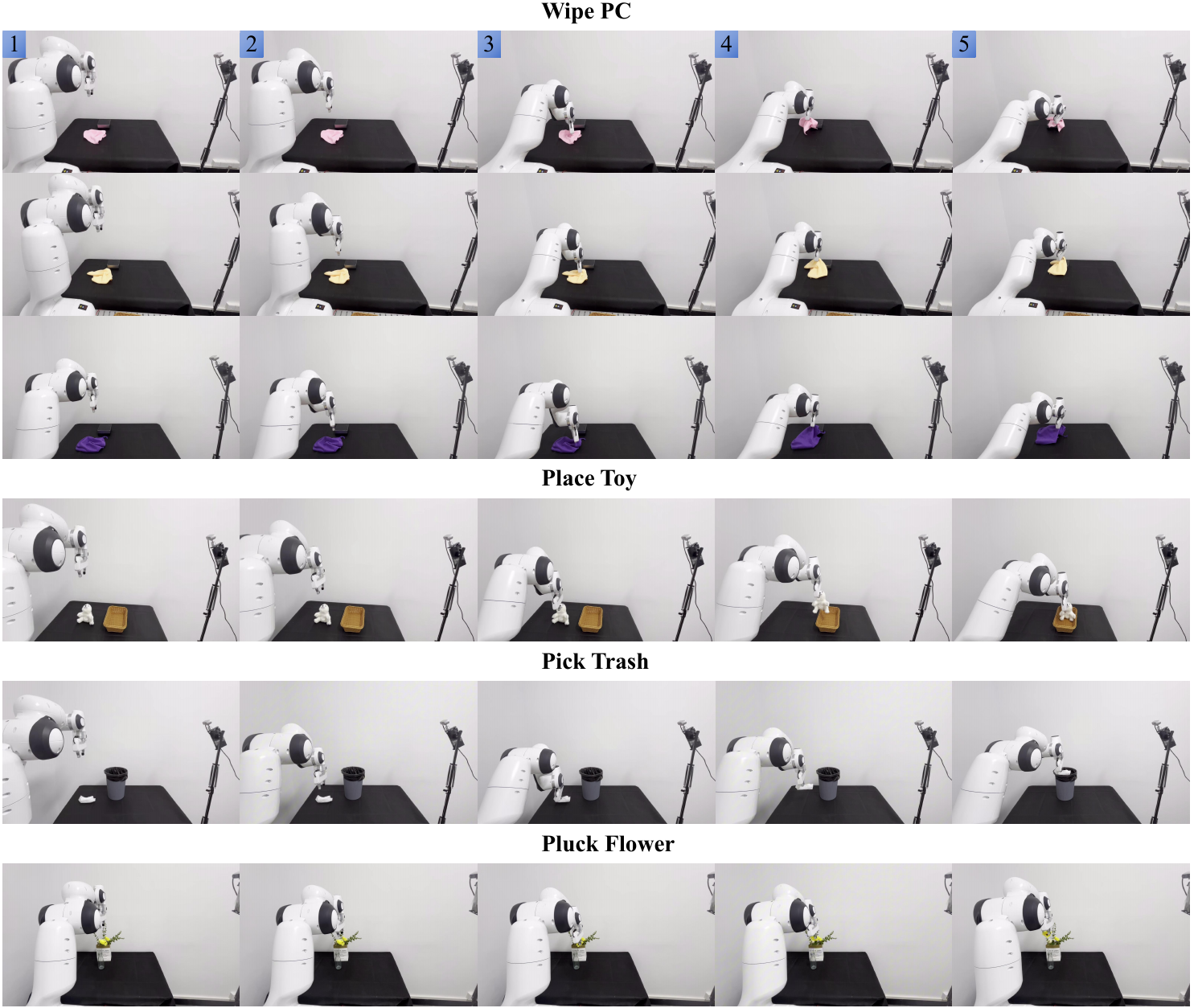}
    \caption{
    Visualization of representative real-world task trajectories.
    The examples show LSA executing real-world manipulation tasks with the Franka Research 3 robot.
    }
    \label{fig:real}
\end{figure}

Fig.~\ref{fig:real} presents representative real-world trajectories executed by LSA.
The examples correspond to the real-world tasks described in the main paper, including \textit{Wipe PC}, \textit{Pick Trash}, \textit{Place Toy}, and \textit{Pluck Flower}.
These visualizations provide qualitative evidence that the learned representation can transfer from human-video pretraining to real-world robot manipulation.

\subsection{More Attentive Guidance Visualizations}
\label{app:more_ag_vis}

\begin{figure}[htbp]
    \centering
    \includegraphics[width=\linewidth]{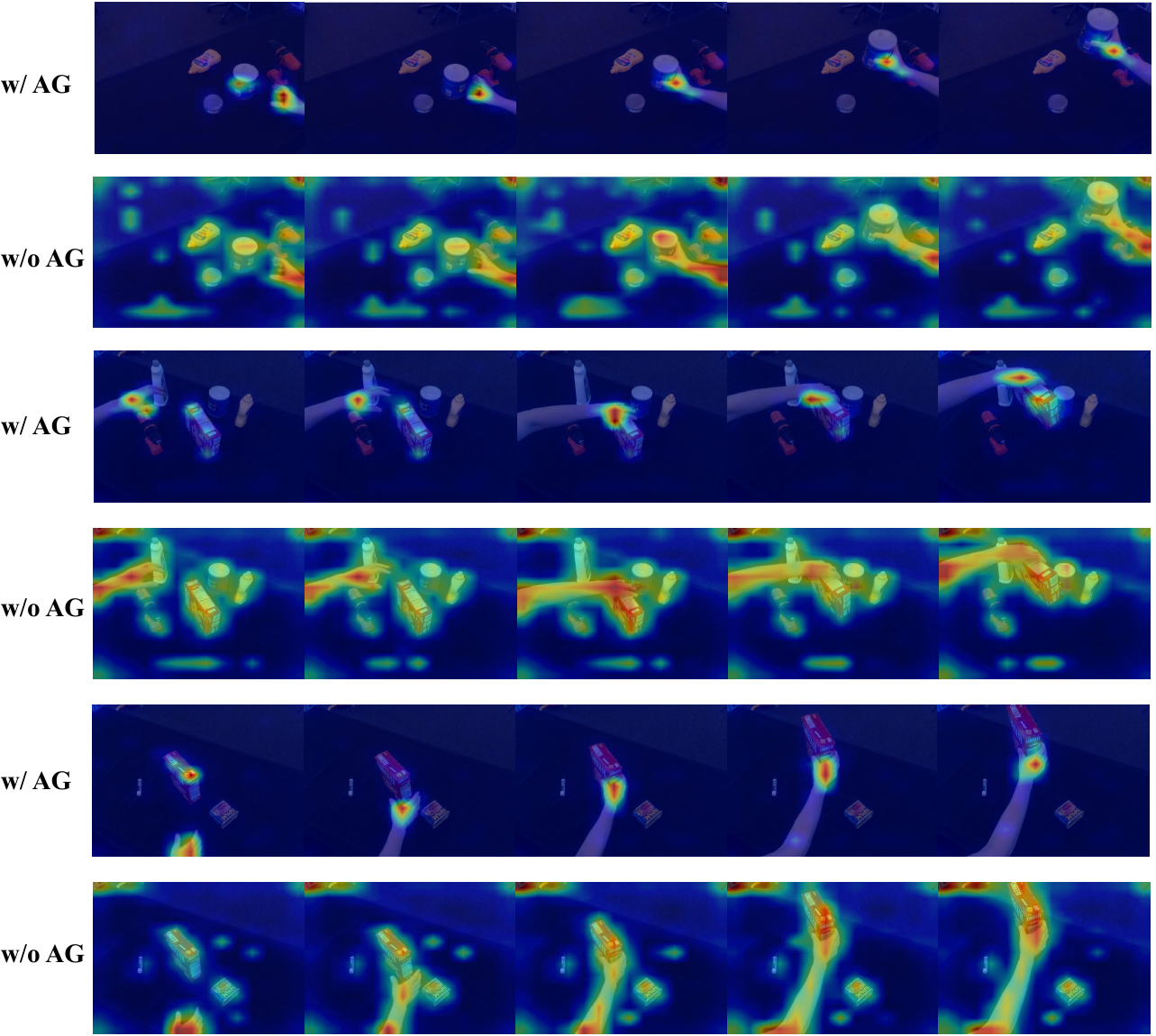}
   
    \caption{
    Additional visualization results of attentive guidance.
    The model learns to focus on hand-centric interaction regions, often covering both the hand and the interacting object.
    }
    \label{fig:att}
    
\end{figure}

\begin{figure}[htbp]
    \centering
    \includegraphics[width=\linewidth]{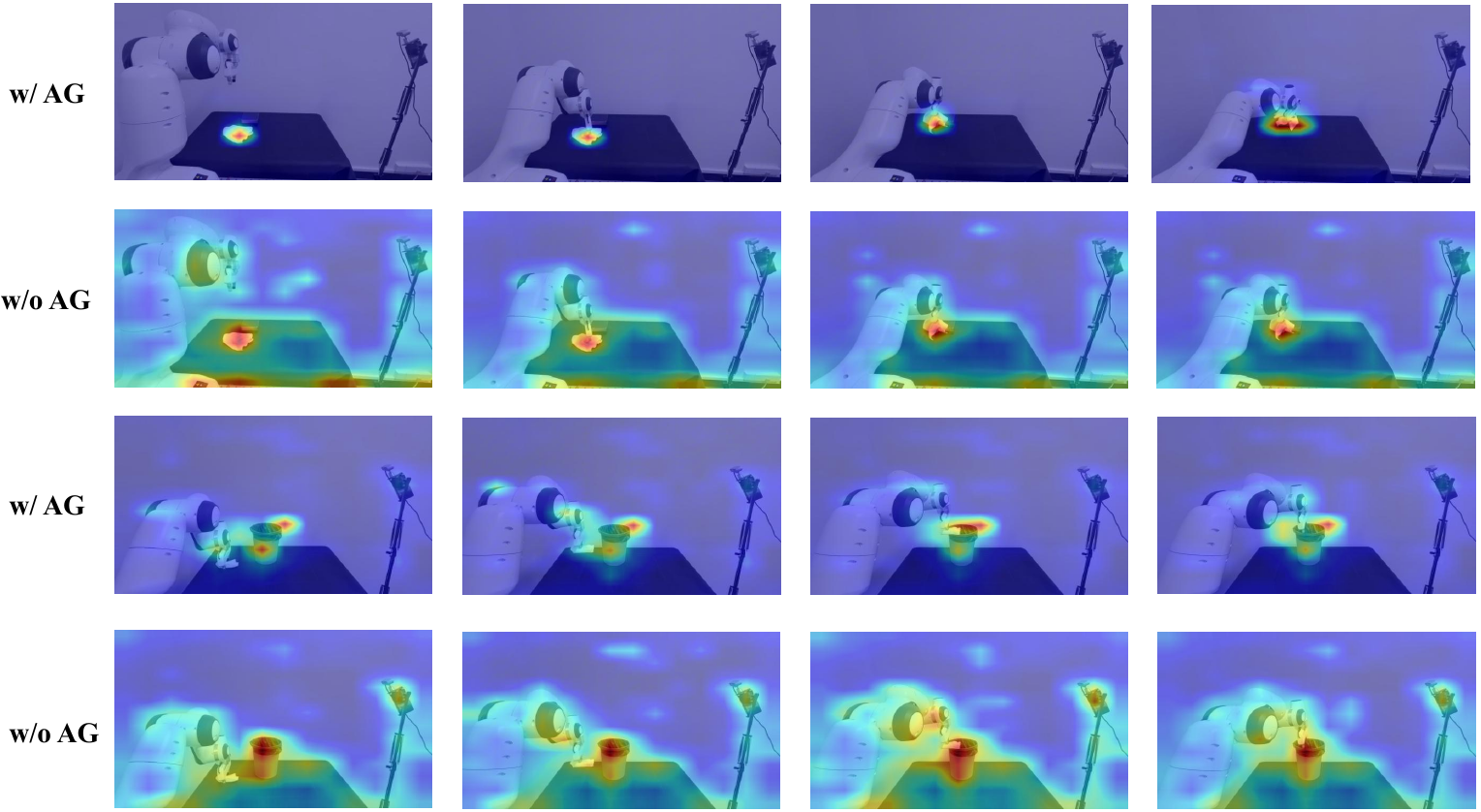}
   
    \caption{
    Additional visualization results of attentive guidance in robotic manipulation scenes.
The model consistently places stronger attention on task-relevant regions, such as the gripper and manipulated objects, while reducing attention to background regions.
    }
    \label{fig:att_robot}
    
\end{figure}

Fig.~\ref{fig:att} presents additional visualization results of attentive guidance on human-object interaction scenes.
The model learns to focus on hand-centric interaction regions, often covering both the hand and the interacting object.
This indicates that the hand prior is not used to impose a fixed hand heatmap, but to mine an internally emerged interaction-aware attention target.
Moreover, Fig.~\ref{fig:att_robot} shows that such attention patterns can transfer to robotic manipulation scenes, where the model places stronger emphasis on task-relevant regions, such as the gripper and manipulated objects, while reducing attention to less relevant background regions.

\newpage
\clearpage
\section*{NeurIPS Paper Checklist}

The checklist is designed to encourage best practices for responsible machine learning research, addressing issues of reproducibility, transparency, research ethics, and societal impact. Do not remove the checklist: {\bf The papers not including the checklist will be desk rejected.} The checklist should follow the references and follow the (optional) supplemental material.  The checklist does NOT count towards the page
limit. 

Please read the checklist guidelines carefully for information on how to answer these questions. For each question in the checklist:
\begin{itemize}
    \item You should answer \answerYes{}, \answerNo{}, or \answerNA{}.
    \item \answerNA{} means either that the question is Not Applicable for that particular paper or the relevant information is Not Available.
    \item Please provide a short (1--2 sentence) justification right after your answer (even for \answerNA). 
\end{itemize}

{\bf The checklist answers are an integral part of your paper submission.} They are visible to the reviewers, area chairs, senior area chairs, and ethics reviewers. You will also be asked to include it (after eventual revisions) with the final version of your paper, and its final version will be published with the paper.

The reviewers of your paper will be asked to use the checklist as one of the factors in their evaluation. While \answerYes{} is generally preferable to \answerNo{}, it is perfectly acceptable to answer \answerNo{} provided a proper justification is given (e.g., error bars are not reported because it would be too computationally expensive'' or ``we were unable to find the license for the dataset we used''). In general, answering \answerNo{} or \answerNA{} is not grounds for rejection. While the questions are phrased in a binary way, we acknowledge that the true answer is often more nuanced, so please just use your best judgment and write a justification to elaborate. All supporting evidence can appear either in the main paper or the supplemental material, provided in appendix. If you answer \answerYes{} to a question, in the justification please point to the section(s) where related material for the question can be found.

IMPORTANT, please:
\begin{itemize}
    \item {\bf Delete this instruction block, but keep the section heading ``NeurIPS Paper Checklist"},
    \item  {\bf Keep the checklist subsection headings, questions/answers and guidelines below.}
    \item {\bf Do not modify the questions and only use the provided macros for your answers}.
\end{itemize}


\begin{enumerate}

\item {\bf Claims}
    \item[] Question: Do the main claims made in the abstract and introduction accurately reflect the paper's contributions and scope?
    \item[] Answer: \answerYes{} 
    \item[] Justification: {The main claims in the abstract and introduction accurately describe the proposed Lift-See-Act framework, including 3D representation alignment, structural guidance, attentive guidance, and motion planning. These claims are supported by the method design in Section 3 and the empirical evaluations reported in Section 4.}

\item {\bf Limitations}
    \item[] Question: Does the paper discuss the limitations of the work performed by the authors?
    \item[] Answer: \answerYes{} 
    \item[] Justification: {We discuss the limitations and future directions of the proposed framework in the conclusion.}

\item {\bf Theory assumptions and proofs}
    \item[] Question: For each theoretical result, does the paper provide the full set of assumptions and a complete (and correct) proof?
    \item[] Answer: \answerNA{} 
    \item[] Justification: {This paper does not present formal theoretical results, theorems, or proofs. The contribution is mainly methodological and empirical, focusing on robot policy pretraining with hierarchical 3D supervision.}

    \item {\bf Experimental result reproducibility}
    \item[] Question: Does the paper fully disclose all the information needed to reproduce the main experimental results of the paper to the extent that it affects the main claims and/or conclusions of the paper (regardless of whether the code and data are provided or not)?
    \item[] Answer: \answerYes{} 
    \item[] Justification: {The paper provides the main algorithmic components, including the data engine, dense representation alignment, depth-map reconstruction, attentive guidance, motion planning objective, and overall loss formulation. The experimental protocol, datasets, metrics, and implementation details are described in Section 4 and the appendix.}

\item {\bf Open access to data and code}
    \item[] Question: Does the paper provide open access to the data and code, with sufficient instructions to faithfully reproduce the main experimental results, as described in supplemental material?
    \item[] Answer: \answerNo{} 
    \item[] Justification: {The code and processed data are not publicly released at the time of submission, although the paper states that the code will be released. We provide detailed methodological and experimental descriptions to support reproducibility before the code release.}

\item {\bf Experimental setting/details}
    \item[] Question: Does the paper specify all the training and test details (e.g., data splits, hyperparameters, how they were chosen, type of optimizer) necessary to understand the results?
    \item[] Answer: \answerYes{}{} 
    \item[] Justification: {The paper specifies the major training and evaluation settings, including datasets, metrics, model components, loss terms, and key hyperparameters. Additional dataset and implementation details are provided in the appendix.}

\item {\bf Experiment statistical significance}
    \item[] Question: Does the paper report error bars suitably and correctly defined or other appropriate information about the statistical significance of the experiments?
    \item[] Answer: \answerYes{} 
    \item[] Justification: {The current version reports the main experimental results and provides error bars. }

\item {\bf Experiments compute resources}
    \item[] Question: For each experiment, does the paper provide sufficient information on the computer resources (type of compute workers, memory, time of execution) needed to reproduce the experiments?
    \item[] Answer: \answerYes{} 
    \item[] Justification: {The current version provides complete compute-resource information, such as GPU type and memory.}
    
\item {\bf Code of ethics}
    \item[] Question: Does the research conducted in the paper conform, in every respect, with the NeurIPS Code of Ethics \url{https://neurips.cc/public/EthicsGuidelines}?
    \item[] Answer: \answerYes{} 
    \item[] Justification: {The research is conducted in accordance with the NeurIPS Code of Ethics. The work uses existing datasets and models for robot policy pretraining and does not involve deceptive data collection, unsafe human-subject experiments, or intentional misuse.}

\item {\bf Broader impacts}
    \item[] Question: Does the paper discuss both potential positive societal impacts and negative societal impacts of the work performed?
    \item[] Answer: \answerYes{} 
    \item[] Justification: {The paper discusses the societal impacts. To the best of our knowledge, this work has no obvious negative social impact.}
    
\item {\bf Safeguards}
    \item[] Question: Does the paper describe safeguards that have been put in place for responsible release of data or models that have a high risk for misuse (e.g., pre-trained language models, image generators, or scraped datasets)?
    \item[] Answer: \answerNA{} 
    \item[] Justification: {The paper does not release high-risk generative models, scraped datasets, or deployed robotic systems that directly enable misuse. The proposed framework is a robot policy pretraining method, and any real-world deployment should still be conducted under controlled safety constraints.}

\item {\bf Licenses for existing assets}
    \item[] Question: Are the creators or original owners of assets (e.g., code, data, models), used in the paper, properly credited and are the license and terms of use explicitly mentioned and properly respected?
    \item[] Answer: \answerYes{} 
    \item[] Justification: {The paper cites the existing datasets, foundation models, and prior codebases used in the study. We follow the corresponding licenses and terms of use for all external assets, and will include explicit license information in the released code and documentation.}

\item {\bf New assets}
    \item[] Question: Are new assets introduced in the paper well documented and is the documentation provided alongside the assets?
    \item[] Answer: \answerNA{} 
    \item[] Justification: {The paper does not introduce a new benchmark dataset or newly collected human-subject dataset. The main contribution is a pretraining framework and associated training objectives built on existing data and models.}

\item {\bf Crowdsourcing and research with human subjects}
    \item[] Question: For crowdsourcing experiments and research with human subjects, does the paper include the full text of instructions given to participants and screenshots, if applicable, as well as details about compensation (if any)? 
    \item[] Answer: \answerNA{} 
    \item[] Justification: {The paper does not conduct new research with human subjects or collect new data from participants.}

\item {\bf Institutional review board (IRB) approvals or equivalent for research with human subjects}
    \item[] Question: Does the paper describe potential risks incurred by study participants, whether such risks were disclosed to the subjects, and whether Institutional Review Board (IRB) approvals (or an equivalent approval/review based on the requirements of your country or institution) were obtained?
    \item[] Answer: \answerNA{} 
    \item[] Justification: {The paper does not conduct new research with human subjects or collect new data from participants. Therefore, IRB approval or equivalent human-subject review is not applicable.}

\item {\bf Declaration of LLM usage}
    \item[] Question: Does the paper describe the usage of LLMs if it is an important, original, or non-standard component of the core methods in this research? Note that if the LLM is used only for writing, editing, or formatting purposes and does \emph{not} impact the core methodology, scientific rigor, or originality of the research, declaration is not required.
    \item[] Answer: \answerNA{} 
    \item[] Justification: {The core method development in this research does not involve LLMs as important, original, or non-standard components.}

\end{enumerate}

\end{document}